\documentclass{article}

\usepackage{microtype}
\usepackage{graphicx}
\usepackage{subfigure}
\usepackage{booktabs}
\usepackage[fleqn]{amsmath}% for professional tables
\usepackage{xcolor}
\usepackage{xspace}
\usepackage[utf8x]{inputenc} 
\usepackage{calligra}
\definecolor{dong}{RGB}{0,0,255}
\definecolor{jw}{RGB}{0,164,164}
\definecolor{jie}{RGB}{0,128,255}
\newcommand{\name}{Flame\xspace}

\usepackage{tikz}
\newcommand*{\circled}[1]{\lower.7ex\hbox{\tikz\draw (0pt, 0pt)%
    circle (.5em) node {\makebox[1em][c]{\small #1}};}}

\usepackage{hyperref}

\usepackage[accepted]{sysml2019}

\sysmltitlerunning{Flame: A Self-Adaptive Auto-Labeling System for Heterogeneous Mobile Processors}

\begin{document}

\twocolumn[
\sysmltitle{Flame: A Self-Adaptive Auto-labeling System for Heterogeneous Mobile Processors}

% unless you've provided the [accepted] option to the sysml2019


\sysmlsetsymbol{equal}{*}

\begin{sysmlauthorlist}
\sysmlauthor{Jie Liu}{equal,ucmerced}
\sysmlauthor{Jiawen Liu}{equal,ucmerced}
\sysmlauthor{Zhen Xie}{ucmerced}
\sysmlauthor{Dong Li}{ucmerced}
\end{sysmlauthorlist}

\sysmlaffiliation{ucmerced}{University of California, Merced.

\textit{MLSys Conference 2020, On-Device Intelligence Workshop},
Austin, TX, USA, 2020.
Copyright 2020 by the author(s)
}

\sysmlcorrespondingauthor{Dong Li}{dli35@ucmerced.edu}

\sysmlkeywords{Machine Learning, SysML}

\vskip 0.3in

\begin{abstract}
How to accurately and efficiently label data on a mobile device is critical for the success of training machine learning models on mobile devices. Auto-labeling data on mobile devices is challenging, because data is usually incrementally generated and there is possibility of having unknown labels. Furthermore, the rich hardware heterogeneity on mobile devices creates challenges on efficiently executing auto-labeling workloads.

In this paper, we introduce \name, an auto-labeling system that can label non-stationary data with unknown labels. \name includes a runtime system that efficiently schedules and executes auto-labeling workloads on heterogeneous mobile processors. Evaluating \name with eight datasets on a smartphone, we demonstrate that \name enables auto-labeling with high labeling accuracy and high performance.   
\end{abstract}
]

%A major open problem in machine learning is how to automatically labeling high quality training data with little effort. In this work, we introduce a new automatically labeling strategy that can labeling the training data in the incremental way as dataset increases: only the limited number of data with label are available at the early stage, while the large number of data without labels, furthermore, both the amount of data and new labels will be dynamically added into the dataset progressively.


% The \sysmlEqualContribution command is standard text for equal contribution.

\printAffiliationsAndNotice{\sysmlEqualContribution} % otherwise use the standard text.

\section{Introduction}
\label{sec:intro}
%({\color{red} Motivate why labeling for the data on the mobile devices is important, first, illustrate why training on the mobile devices; second, describe that lack of labeled data is the essential problem for training on mobile devices. Thus, the problem about labeling the data on mobile devices is meaningful}) The success of machine learning (ML) on mobile devices has ignited the interest in using ML for a variety of tasks. 
Machine learning (ML) has been increasingly utilized in mobile devices (e.g., face and voice recognition and smart keyboard). However, many ML applications running on mobile devices (such as smartphone and smart home hub) mainly focus on ML inference not ML training. 
%It means the parameters of the machine learning models trained on server side and the related configure files are frequently transformed from the server side to the mobile devices. Such frequent data transportation may bring large bandwidth overhead both for the server and mobile devices; furthermore, the model trained on the server side cannot personalize well for the data generated on the mobile devices which decreases the user experience. Based on these limitations of inference,
Recently, training ML models on mobile devices attract attentions because of the concerns on data privacy, security, network bandwidth, and availability of public cloud for model training~\cite{eom2015malmos, konevcny2016federated}. Training ML models leverages data generated locally in mobile devices without uploading data to any public domain. However, data generated on local mobile devices usually does not have labels, causing difficulty for training ML models (especially supervised ML models). For example, the user uses a smartphone to take pictures. Those pictures are seldom labeled, and are difficult to be used to train ML models. How to accurately and efficiently label data on a mobile device is critical for the success of training ML models on the mobile device. 

%training machine learning models on a mobile device with the data generated by the users has attracted more and more attentions among the researchers, like federated learning. However, the data generated on the mobile devices are lack of labels, this problem has impeded bringing AI for mobile devices. Thus, how to accurately and efficiently label for the data generated on the mobile devices is an essential issue to be addressed. 

%({\color{red} The characteristics of data generated on mobile devices}) 
Using automatic labeling is a solution to address the above problem. Studies of automatic labeling in the past decade have been focusing on data stored on servers~\cite{Varma2018SnubaAW, ratner2017snorkel, yang2018cost, haas2015clamshell}. Such data usually has a fixed size, and labels are pre-determined and fixed. %\textcolor{jie}{These proposed approaches usually rely on techniques like weak supervision, crowdsourcing, boosting, or user-defined labeling functions to assign noisy labels for unlabeled data.} 
However, the data generated on a mobile device have different characteristics, compared with the data located on a server. 
%First, data on mobile devices may be dynamically generated during the usage, and the new data may have new labels that are never known before. Second, labeling efficiency (i.e., execution time and energy consumption) is a big concern on mobile devices, because of its resource constraint. 
%({\color{red} Drawbacks of the existing work when applied into labeling for data on mobile devices}) 
In particular, on a mobile device, data is incrementally added. For example, pictures are added into a smartphone on a daily base, as the user takes pictures from time to time. Also, as data is dynamically generated, some new labels may appear. However, most of the existing methods cannot recognize new labels. %%Their labeling quality can be drastically reduced, because of the occurrence of data with unknown labels. %Also, there could be multiple labels, not just two labels as in the existing work \textcolor{dong}{[xxx]}. 

%Studies of automatic labeling in the past decade have been focusing on data stored on servers. Such data usually has a fixed size and a fixed set of labels. These proposed approaches usually rely on weak supervision, or methods that can assign noisy training labels to unlabeled data, like distant supervision, crowdsourcing, boosting, and user-defined labeling functions. However, the data generated on the mobile devices have different characteristics compared with the data located on the server side, e.g, new data on mobile devices may dynamically generated during the usage, these new data may have new labels which never seen before, efficiency is a big concern for mobile devices due to its resource constraint, etc.

%({\color{red} The characteristics of data generated on mobile devices}) Studies of automatic labeling in the past decade have focused on data stored on the server side, such data always with fixed size and fixed set of labels. These proposed approaches usually rely on weak supervision, or methods that can assign noisy training labels to unlabeled data, like distant supervision, crowdsourcing, boosting, and user-defined labeling functions. However, the data generated on the mobile devices have different characteristics compared with the data located on the server side, e.g, new data on mobile devices may dynamically generated during the usage, these new data may have new labels which never seen before, efficiency is a big concern for mobile devices due to its resource constraint, etc.

Besides distinguished data characteristics, auto-labeling on mobile devices face a challenge on hardware heterogeneity on mobile devices. Mobile devices are often equipped with mobile processors with rich heterogeneity for high energy efficiency and performance~\cite{wu2019machine,DBLP:journals/corr/abs-1906-04278}. For example, Samsung S9, a mobile device we study in this paper, has two types of CPU cores and a mobile GPU. Scheduling computation  becomes complicated as we have hardware heterogeneity, because we must decide where to run computation and how to control thread-level parallelism for short execution time and low energy consumption. Leveraging heterogeneous mobile processors to efficiently execute the auto-labeling workload is a key to enable feasible auto-labeling on mobile devices. 

%({\color{red} Drawbacks of the existing work when applied into labeling for data on mobile devices}) Current existing labeling methods mainly focus on static dataset with fixed number of labels, they are not suitable when data dynamically increment. However, such scenario is quite common for labeling data located on mobile devices. Furthermore, some new labels may appear as data dynamically generated. Thus, the labeling task should have the capability to address the data with multiple labels, while the existing work mainly focus on the task related with binary classification. Moreover, current methods can not recognize the appeared novel labels, so their labeling quality may drastically reduce due to occurrence of data that may not belong to the known label set.

%({\color{red} Illustrate the limitations faced by the existing work have been solved by our system}) 
In this paper, we introduce an auto-labeling system, \textit{\name}, in order to address the above challenges. \name is particularly designed for mobile devices with heterogeneous mobile processors. \name is featured with self-adaptiveness to handle incrementally generated data with unknown labels. In particular, \name uses a cluster based technique to gather and detect data that belongs to the same class but with new features. \name then creates and assigns new labels without using the existing labels.

Furthermore, \name is featured with a hardware heterogeneity-aware runtime system. To efficiently schedule computation kernels, the runtime system profiles performance of kernels on different computing units, based on high predictability of the auto-labeling workload. Using the profiling results, the runtime system performs kernel scheduling based on a set of greedy heuristic policies. To efficiently run individual kernels, the runtime system chooses optimal number of threads (i.e., thread-level concurrency control) and divides the kernel computation between heterogeneous mobile processors, using a couple of analytical performance models.

%({\color{red} Briefly introduce our mobile runtime system design}) To address the above challenges, we introduce a runtime system with heterogeneity-aware inter- and intra-kernel parallelism. To optimize the hardware efficiency and address the straggler effect across kernels, we first profile and characterize kernels which are automatically processed online. We then perform heterogeneity-aware kernel scheduling using several greedy-heuristic strategies based on the kernel profiling and features of mobile processors. To address the parallelism issue of load imbalance within a kernel, we propose a heterogeneity-aware intra-kernel division based on two analytical prediction models to dynamically divide the parallel kernel for heterogeneous mobile processors.  

%({\color{red} Our contributions})We describe $Flame$, a self-adaptive auto-labeling system to assign training labels to a large, unlabeled dataset on mobile devices using a small labeled dataset. A summary of our contributions are as follows:

We summarize our contributions as follows. 

\begin{itemize}
    \item We propose a self-adaptive auto-labeling system to assign labels for data on mobile devices. To our best knowledge, this is the first auto-labeling system focusing on the data labeling problem on mobile devices.
    
    \item We present a new labeling algorithm to process dynamically generated data (i.e., non-stationary data) with unknown labels.
    
    \item We introduce a runtime system to efficiently leverage heterogeneous mobile processors for auto-labeling. We evaluate our system on a smartphone, and demonstrate high labeling quality and high performance with \name. 
\end{itemize}

%$1)$ We propose a self-adaptive auto-labeling system to assign labels for the data on mobile devices, to our best knowledge this is the first auto-labeling system focuses on the labeling problem for the data located on the mobile devices.

%$2)$ We present a new labeling strategy which is suitable for the data dynamic increases and novel labels may appear over the non-stationary data. Such labeling strategy is different from the existing work as they can only label for the data with fixed size and fixed set of labels. 

%$3)$ To improve the efficiency of executing the auto-labeling system on mobile devices, we propose a runtime system equipped with hetero-geneity-aware inter- and intra-kernel parallelism approaches to highly improve the system throughput and the utilization of heterogeneous mobile processors. 
\section{Background}
\label{sec:background}

\begin{figure*}
	\centering
	\includegraphics[width=1\linewidth]{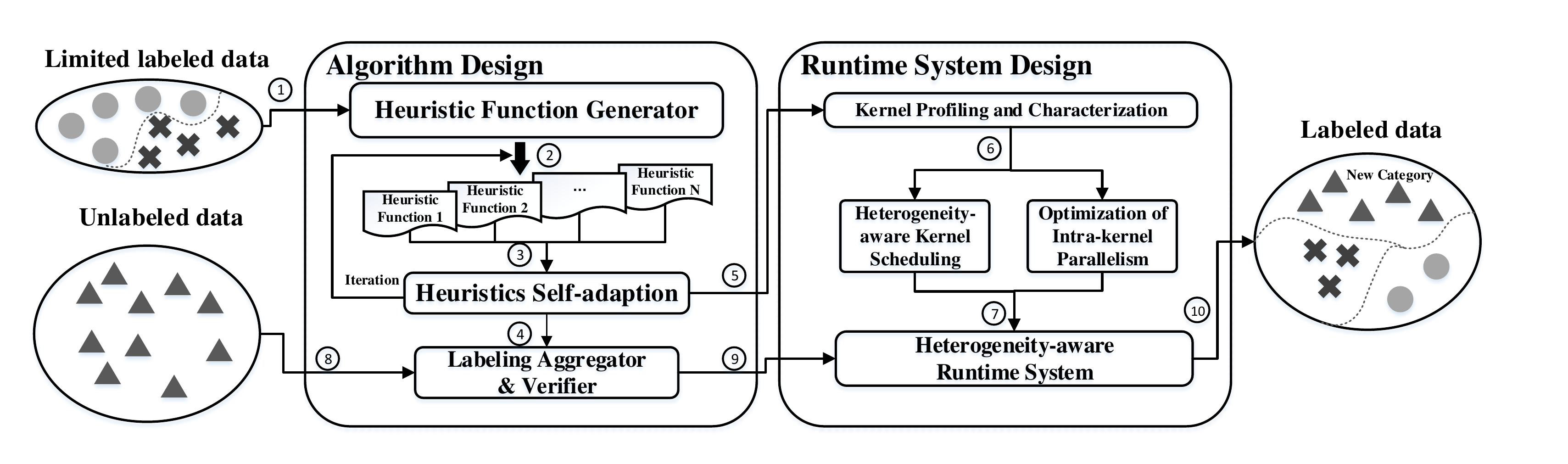}
	%\vspace{-18pt} 
	\caption{Overview of the Flame system.} \centering
	%\vspace{-12pt} 
	\label{fig:overview} 
\end{figure*}

\subsection{Auto-labeling and Preliminaries}

An important problem in ML is how to automatically label a relatively large quantity of unlabeled data with little labeled data \cite{mei2007automatic, mao2012automatic, dube2019automatic}. When the user gets a new mobile device, there is usually a limited number of labeled data. The goal of our auto-labeling system is to label gradually increasing unlabeled data. An existing solution to address the traditional auto-labeling problem on servers is called co-training \cite{qiao2018deep}. In this solution, the label of each data can be determined with two conditionally independent labeling functions by utilizing both labeled and unlabeled data. 
Another existing solution on servers uses the boosting method \cite{Varma2018SnubaAW}, which constructs and combines many ``weak'' classifiers into a ``strong'' one. However, for the auto-labeling problem on mobile device, co-training is difficult to derive precise classifiers, because of the insufficiency of labeled data; The boosting method cannot work either, because it assumes a fixed number of labels, and hence cannot label an unfixed number of labels. In our paper, we combine the co-training and boosting methods to deal with the auto-labeling problem on mobile devices. 
%Furthermore, the data to be labeled can be divided into some small partitions that can be distinguished from the total data by a trained classifier. Therefore, in our paper, we will combine the co-training and boosting methods to deal with the dynamic problem with high precision.

%An important problem in machine learning is how to automatically labeling relatively large quantity of unlabeled data with little labeled training data \cite{XXX}. Specifically, the auto-labeling problem only have limited number of labeled data when user just get a new mobile device, and its goal is to label gradually increasing unlabeled data. One solution for auto-labeling problem is called co-training \cite{XXX}, in which the label of each instance can be partitioned into two conditionally independent views by utilizing both of the labeled and unlabeled data. Another solution used for auto-labeling is boosting \cite{XXX}, which constructs and combines many “weak” classifiers into a "strong" one. However, for auto-labeling problem on mobile device, co-training is difficult to derive precise classifiers by insufficient labeled data, and boosting is based on a fixed number of categories that result in inability to labeling dynamically changing number of categories. Furthermore, the data need to be labeled can be divided into some small partitions that can be distinguished from the total data by a trained classifier. Therefore, in our paper, we will combine the co-training and boosting methods to deal with the dynamic problem with high precision.

We define several terms used in the paper as follows. The \textit{prototype} indicates an average or best exemplar of a category (label), which could represent the instances of an entire category. A prototype is usually denoted as a tuple $p$, each element in the tuple represents some important patterns and characteristics of the category.

The \textit{heuristic function} in our paper is an union of several prototypes, denoted as $p_{1}\cup p_{2} \cup ... \cup p_{t}$, where $t$ is the number of prototypes contained by the heuristic function. Updating each heuristic function is through updating its prototypes. The process for maintaining the optimal prototypes for a category is called \textit{prototype learning}. In our study, prototypes are learned through optimizing a self-defined object function. We combine the prototype learning with the boosting method for auto-labeling, which generates more discriminating and robust results.  

%Some preliminaries of this paper are as follow: "Prototype" indicates an average or best exemplar of a category, which could represent the instances of the entire category. A prototype is usually denoted as a tuple $p$, each element in the tuple represents some important patterns and characteristics with same category.

%\textcolor{dong}{(please double check if the following two paragraphs are correct.)}
%The ``feature'' of a data sample includes interesting points that are calculated by quadratic localization, their scales, a repeatable angle that is extracted for each interest point, and so on. Based on the features, we can build prototypes that present some fragments of data with high probability.

%"Feature" represents including interest points that are calculated by quadratic localization, their scales, repeatable angle that is extracted for each interest point, and so on. Based on these features, we can build prototypes that present some fragments of data with high probability.

%In our works, prototypes are learned through optimizing the self-defined object function. Furthermore, we also combine the prototype learning with boosting method for auto-labeling task, which could predict more discriminating and robust result.  

\subsection{Heterogeneous Mobile Processors}
The System on Chips (SoCs) in mobile devices increasingly employs heterogeneous mobile processors, such as CPU, GPU, DSP, and NPU. Samsung S9, a mobile device we used for study, uses Qualcomm 845 SoC, which includes a 4-core fast CPU, a 4-core slow CPU, and an Adreno 630 mobile GPU. The fast and slow CPU cores are different in terms of frequency, cache hierarchy, instruction scheduling and energy efficiency. The mobile GPU is particularly efficient for processing data-intensive tasks. 
%\textcolor{dong}{Please complete the above paragraph.}

Figure~\ref{fig:overview} gives an overview of \name, including the algorithm design described in Section~\ref{sec:model} and system design described in Section~\ref{sec:sys}. The circled numbers in Figure~\ref{fig:overview} depicts the whole workflow, which are described in the following two sections.  %\textcolor{green}{(it maybe can be placed in a better place.)}

\section{Model Design}
\label{sec:model}
%In this part, we briefly describe the input and output of the system; besides, we also introduce some important notations that will be used in the rest of the paper.
In this section, we describe the auto-labeling algorithm design for \name. As shown in the left part of  Figure~\ref{fig:overview}, the algorithm design includes three components: a heuristic function generator to generate a number of heuristic functions for assigning labels, a self-adaptive mechanism for updating the heuristic functions and detecting whether an unknown label appears, and a labeling aggregator for combining and verifying the confidence of label assignments. We describe the three components and input/output of \name as follows.

%As the left part of Figure~\ref{fig:overview} shows, three components in the algorithm design are a number of heuristic functions generated by a heuristic function generator, the heuristics self-adaptive mechanism for updating the heuristic functions and detecting whether an unknown label appears, finally, the component related with labeling aggregator is for combining and verifying the confidence of the assigned label.

\subsection{Input and Output Data}
\textbf{Input Data.} The input data of \name is a small number of data with labels and a large number of data without labels. Each data is defined by its primitives. In many of our labeling use cases, the primitives can be viewed as the basic features associated with the corresponding data. For instance, in the use case of labeling, primitives can be color, size, shape, etc. In our work, we want to label the data in an automatic way, so we extract the features of the images as the primitives in our auto-labeling algorithm. Given a labeled dataset $D_L = \{x_i, y_i\}^{N_L}_{i=1}$, where $x_i \in R^{d}$ is the primitives of the data $i$ and $y_i \in Y = \{1, 2,..., C\}$ is the associated true label, $R^{d}$ represents a $d$ dimensional space for data and $C$ is the total number of labels in $Y$. The non-stationary unlabeled dataset $D_U = \{x_t\}^{N_U}_{t=1}$, where $x_t \in R^{d}$ and $N_U \in [0, \infty)$ represents the number of the unlabeled data. In our setting $N_U$ can be very large as the new data is dynamically generated.

%\textbf{Input Data.} The input data for the system is a small number of data with labels and a large number of data without labels. Each data is defined by its corresponding primitives, the primitives can be viewed as the basic features associated with the corresponding data. For instance, in the task of labeling the category for the different flowers, primitives can be color, size, shape, etc. In our work, we want to label the data in an automatic way, so we extract the features of the images as the primitives in our algorithm. Given the limited number of labeled dataset $D_L = \{x_i, y_i\}^{N_L}_{i=1}$, where $x_i \in R^{d}$ is the primitives of the data and $y_i \in Y = \{1, 2,..., C\}$ is the associated true label. The non-stationary unlabeled dataset $D_U = \{x_t\}^{N_U}_{t=1}$, where $x_t \in R^{d}$ and $N_U \in [0, \infty)$ represents the number of the unlabeled data. In our setting $N_U$ can be very large as the new data dynamically generated.

\textbf{Output Data.} The output of \name is the confidence of a label $y_i \in Y^{'} = \{1, 2,..., C, ..., C^{'}\}$ for data $i$ in the unlabeled dataset $D_U$ ($Y^{i}$ is the set of result labels including old labels and detected new labels). Here, $C^{'} \geq C$, which indicates that some new labels that are not in $Y$ may appear in $Y^{'}$, as new data is incrementally generated. The confidence value is calculated through an ensemble method in \name, discussed in Section~\ref{sec:labeling_aggregator}. 

%\textbf{Output Data.} The output of the system is the confidence of the label $y_t \in Y^{'} = \{1, 2,..., C, ..., C^{'}\}$ for data in the unlabeled dataset $D_U$. Here, $C^{'} \geq C$ means some novel labels which are not in the $Y$ may appear as new data incrementally generated. The confidence value is calculated through an ensemble method in the system, it combines the confidence value generated by each heuristic function to improve the overall performance.

%\subsection{Overview}
%The main idea of this paper is to automatic labeling for the unlabeled data such that: $1)$ the labels of  unlabeled data in $Y$ can be correctly assigned; $2)$ a novel label $y^{*} \in Y^{'}-Y$, where  $y^{*} \in Y{'} \wedge y^{*} \notin Y$, can be immediately detected, and $3)$ heuristic functions for data labeling can be self-adapted over a non-stationary unlabeled data. As figure 1 shows, three components in the system are the heuristic function pool that contains all the heuristic functions for labeling, incremental learning for detecting whether the novel label $y^{*}$ appears and self-adaptation over the non-stationary unlabeled dataset, and an aggregator for combining the confidence of the assigned label. 

\subsection{Heuristic Functions Generation}
Existing studies~\cite{ratner2017snorkel, yang2018cost} on heuristic function generation for auto-labeling have demonstrated the success of using machine learning models (e.g., Decision Tree, Logistic Regression, K-Nearest Neighbor) as the heuristic functions. 
%A large number of heuristic functions will be generated for each possible combinations of 1 to M primitives, resulting in at most $\sum_{M^{'}=1}^{M}\binom{M}{M^{'}} = 2^{M}-1$ heuristic functions. 
However, these methods can be costly because of two reasons. (1) A large number of heuristic functions are generated, which bring high computation overhead during the labeling, especially on mobile devices with limited computation resources; (2) Each heuristic function has parameters. Updating parameters from all heuristic functions demands large amount of computation. Therefore, we design a heuristic function generation method to address the above problems, while being cognizant of computational and memory accessing efficiency required for \name.

%%Existing studies which address the problem of heuristic function generation have demonstrated the use of machine learning models as the heuristic functions, e.g., Decision Tree, Logistic Regression, K-Nearest Neighbor, etc. 
%A large number of heuristic functions will be generated for each possible combinations of 1 to M primitives, resulting in at most $\sum_{M^{'}=1}^{M}\binom{M}{M^{'}} = 2^{M}-1$ heuristic functions. 
%%The primary disadvantage of such methods is too many heuristic functions will be generated which brings too much computation overhead during the labeling, especially when the computation resources are limited. Moreover, too many heuristic functions bring too much parameters, while updating these parameters needs large amount of computation. Therefore, we design a heuristic function generation method to address these concerns while being cognizant of the computational and memory accessing efficiency required for the system. 

Each heuristic function works well for a part of data in the primitive space. We expect that the heuristics in the system together have high data coverage. The larger the coverage is, the higher the labeling cost could be reduced. In \name, we use a clustering algorithm to determine the boundary of each heuristic function, because of the flexibility of the clustering algorithm. In particular, each heuristic function consists of several clusters generated through an impurity based K-means algorithm based on the initial limited number of labeled data points $D_L = \{x_i, y_i\}^{N_L}_{i=1}$. A cluster is pure if it contains labeled data points from only one class (along with some unlabeled data). Once the clusters are created, the raw data points are discarded to save memory space. %for both computation and memory efficiency. 
The discarded data points in a cluster are replaced by a prototype of the cluster. A prototype of a cluster is a tuple denoted by $p=<c,r,\mu,s,\bar{f}>$, where $c$ is the centroid of the cluster, $r$ represents the radius which is the distance between the centroid and the farthest data point in the cluster, $\mu$ is the mean distance between each data point and the centroid in the cluster, $s$ is the total number of data points in the cluster, and $\bar{f}$ is a vector recording the number of data points belonging to different labels (referred as \textit{frequencies} in the rest of the paper). For example, $\bar{f} = (f_{1}, f_{2}, ..., f_{t})$, where each element $f_{i}$ in $\bar{f}$ is the frequency of the corresponding label $y_{i}$ assigned to the existing data. Each heuristic function $h$ is a collection of K prototypes, $h=\{p_1,...,p_k\}$. 

%Each heuristic function works well for a part of data in the primitive space. For each function, these data are located in the boundary of the function. Then, the heuristics in the system together will have high coverage. Intuitively, the larger the coverage is, the higher the cost on labeling could be reduced. Here, we appeal to use the clustering algorithms to determine the boundary of each heuristic function for its flexibility. Each heuristic function is consisted by several clusters generated through an impurity based K-means algorithm based on the initial limited number of labeled data points $D_L = \{x_i, y_i\}^{N_L}_{i=1}$. A cluster is pure if it contains labeled data points from only one class(along with some unlabeled data). Once the clusters are created, the raw data points are discarded for both computation and memory efficiency. Here, the discarded data points in one cluster will be replaced by a prototype of the corresponding cluster. A prototype of a cluster is a tuple denoted by $p=<c,r,\mu,s,f>$, where $c$ is the centroid of a cluster, $r$ represents the radius which equals to the distance between the centroid and the farthest data point in the corresponding cluster, $\mu$ is the mean distance between the data point and the centroid in a cluster, $s$ is the total number of data points in a cluster, $f$ is a vector records the number of data points belonging to the different labels(referred to as frequencies). Therefore, each heuristic function is a collection of K prototypes, $h=\{p_1,...,p_k\}$. 

When building the heuristic functions using the impurity K-means algorithm, the objective is to minimize the dispersion and impurity of data points contained in each prototype. Thus, the objective function for building the heuristic function is formulated as follows.
\begin{equation}
\label{eq:loss_eq}
Loss(h) = Loss(K-means) + \lambda \ast Loss(Impurity)
\end{equation}
%\begin{equation}
%Loss(\mathbcal{h}) = \sum_{i=1}^{K}\sum_{x \in \mathbcal{D_{p_{i}}}}{\lVert x-\mu_{i} \rVert}^{2} + \sum_{i=1}^{K}\mathbcal{\lambda_{i}}\ast Impurity_{i}
%\end{equation}

In Equation~\ref{eq:loss_eq}, $Loss(K-means)$ is the loss value caused by the dispersion of data points contained in each prototype.
$Loss(K-means)$ is calculated with $\sum_{i=1}^{K}\sum_{x \in %\mathbcal
{D_{p_{i}}}}{\lVert x-\mu_{i} \rVert}^{2}$, where $K$ is the total number of prototypes in $h$ and ${D_{p_{i}}}$ is the set of all data points in the prototype $p_{i}$.
In Equation~\ref{eq:loss_eq}, $Loss(Impurity)$ is the loss value caused by the impurity of data points in each prototype, and $\lambda$ is a hyper-parameter controlling the importance of $Loss(Impurity)$. $Loss(Impurity)$ is calculated as follows: $Loss(Impurity) = Label\_diverse * Entropy$, where $Label\_diverse$ quantifies labeling diversity in a prototype and $Entropy$ is the entropy value of data points in the prototype. A small $Label\_diverse$ leads to small impurity.
$Label\_diverse$ is calculated based on each data point's labeling dissimilarity in the prototype (particularly, $Label\_diverse = \sum_{x \in {D_{p}}}LD(x,y)$, where LD(x,y) of a data $x$ in the prototype with the label $y$ is the total number of labeled points in the prototype that should have labels other than $y$. $LD(x,y) = 0$, if a data point is unlabeled;
Otherwise, $LD(x,y) = \lvert {L}-{L_{\ell}} \rvert$, when the data point $x$ is labeled and its label $y=\ell$, where ${L}$ and ${L_\ell}$ are the sets of all labeled data points and labeled data points with the label $\ell$ in the prototype, respectively.
Thus, $Label\_diverse$ can be written as $\sum_{x \in {D_{p}}}\lvert {L}-{L_{\ell}} \rvert$.
Furthermore, $Entropy = \sum_{\ell = 1}^{C^{'}}(-p_{\ell}*\log(p_{\ell}))$, where $p_{\ell}$ is the  probability of labeling $\ell$ ($p_{\ell} = \frac{\lvert L_{\ell} \rvert}{\lvert L \rvert}$). 

Based on the above discussion, the loss function can be re-formulated as follows.
\begin{small} 
\begin{align}
\!\!\!\!\!\!\!\!\!\!\!\!Loss(h)\!
&=\! \sum_{i=1}^{K}\sum_{x \in D_{p_{i}}}{\lVert x-\mu_{i} \rVert}^{2} + \sum_{i=1}^{K}\lambda_{i}*\sum_{x \in D_{p_{i}}}\lvert L_{i}-L_{i}(\ell) \rvert \nonumber\\
&\quad\, *\sum_{\ell = 1}^{C^{'}}(-\frac{\lvert L_{i}(\ell) \rvert}{\lvert L_{i} \rvert}*\log(\frac{\lvert L_{i}(\ell) \rvert}{\lvert L_{i} \rvert}))  
\end{align}
\end{small} 

Minimizing $Loss({h})$ can get the optimal prototypes for each heuristic function. Each prototype corresponds to a ``hypersphere'' in the primitive space with a centroid and radius. The coverage of a heuristic function $h_i$ is the union of the hyperspheres encompassed by all prototypes in $h_i$. The coverage boundary of the heuristic function pool is the union of coverage of all heuristic functions $h_i \in H$. If a data point $x$ is inside the coverage boundary of $H$, it is labeled using each $h_i \in H, i \in 1...t$ as follows. Let $p_{j}$ is the prototype of $h_i$ whose centroid is the closest to $x$. In the prototype $p_j$, assume $f_{max}$ is the highest frequency value in the frequency vector $\bar{f}$, then $f_{max}$'s corresponding label $y$ will be assigned to the unlabeled data point $x$. Each $h_i \in H$ maintains an assigned label and this label's confidence value for data point $x$. %Finally, the data point $x$ is labeled using the heuristic function set $H$ by taking the majority vote among all the heuristic functions.
Finally, the label for the data point $x$ is determined by taking the majority vote among all heuristic functions.

%Minimizing $Loss(\mathbcal{h})$ can get the optimal prototypes for each heuristic function. Each prototype corresponds to a ``hypersphere'' in the primitive space with a centroid and radius. The coverage of a heuristic function $h_i$ is the union of the \textcolor{blue}{hyperspheres} encompassed by all prototypes in $h_i$. The heuristic function pool in the system may contain $t$ heuristic functions. Thus, the coverage boundary of the heuristic function pool is the union of coverage of all heuristic functions $h_i \in H$. If a data point $x$ is inside the coverage boundary of $H$, it is labeled using each $h_i \in H, i \in 1...t$ as follows. Let $p_j \in h_i$ be the prototype whose centroid is the nearest from $x$. The label of $x$ is assigned as the label that has the highest frequency in the vector $f$ of prototype $p_j$. Finally, the data point $x$ is labeled using the heuristic function set $H$ by taking the majority vote among all the heuristic functions.

\subsection{Labeling Heuristics Self-adaptation}
Many automatic labeling methods~\cite{dunnmon2019cross, ratner2017snorkel, yang2018cost, varma2019learning} assume that the number of possible labels associated with data points is known and fixed. However, in some cases, this is not true. Data points belonging to unknown labels may appear as the dataset dynamically increases. 

In \name, if a data point $x$ is outside of the coverage boundary of $H$, $x$ is regarded as a data point with an unknown label and stored in a buffer $B$. This buffer is periodically checked to observe whether there are enough data points in the buffer with the same new label. We use a distance based method called q-Neighborhood Silhouette Coefficient~\cite{masud2010classification}, shorted as q-NSC to address the problem about detecting new labels. $q$ is a predefined parameter. q-NSC considers both cohesion and separation of data points located in the primitive space, and yields a value in $[-1,1]$. A positive q-NSC value of the data point $x$ indicates that $x$ is close to other $q$ data points in the buffer $B$. This means that these $q$ data points together may have the same potential unknown label. %%$y^{*} \in Y^{'}-Y$, such that $y^{*} \in Y{'} \wedge y^{*} \notin Y$.

%Traditionally, automatic labeling task assumes the number of possible labels associated with the data points is known and fixed. However, in some cases, this is not true. Data points belonging to unknown labels may appear as the dataset dynamically increases. Here, if a data point $x$ is outside of the coverage boundary of $H$, $x$ can be potentially regarded as a data point with unknown labels and stored in a buffer $\mathcal{B}$. This buffer is periodically checked to observe whether there are enough data points in the buffer with the same novel label. Here, we use a distance based method called q-Neighborhood Silhouette Coefficient, shorted as q-NSC\cite{} to address this problem, $q$ is a predefined parameter. q-NSC considers both cohesion and separation of the data points located in the primitive space, and yields a value between -1 and +1. A positive q-NSC value of data point $x$ indicates that $x$ is close to other $q$ data points in the buffer $\mathcal{B}$. It means these $q$ data points together may have the same potential unknown label $y^{*} \in Y^{'}-Y$, such that $y^{*} \in Y{'} \wedge y^{*} \notin Y$.

Because of the dynamic characteristic of the dataset and the requirement of continuous updating prototypes for existing heuristic functions, \name must have an ability to adapt to the changes over the non-stationary dataset without increasing memory footprint and computing overhead. We introduce a mechanism to incrementally incorporate new label information from data points in the buffer $B$ to the existing heuristic functions without loosing discriminatory characteristics. %Here, we utilize the prototype initialization and update to construct the novel heuristic functions. 
Moreover, to limit the memory usage, we set the maximum number of prototypes across all heuristic functions to be $M$.
%Here, we resort to the use of a fixed-size buffer (denoted by $\mathcal{B}$) to store the data point with potential novel labels.

%Here, due to the dynamic characteristic of the dataset and the requirement of continuous updating prototypes for existing heuristic functions. Our system should have the ability to adapt to the changes over the non-stationary dataset without increasing its memory footprint and computing overhead. In this paper, we develop a mechanism to incrementally incorporate novel label information from data point in the buffer $\mathcal{B}$ to the existing heuristic functions without loosing discriminatory characteristics. %Here, we utilize the prototype initialization and update to construct the novel heuristic functions.  Moreover, to limit the memory usage, we fix the maximum number of prototypes across all heuristic functions to be $\mathcal{M}$.
%Here, we resort to the use of a fixed-size buffer (denoted by $\mathcal{B}$) to store the data point with potential novel labels.

Algorithm 1 depicts the mechanism, including self-adaptation of heuristic functions and corresponding updates on  prototypes. \name periodically checks the buffer $B$ and requests checking on potential new label. After a new label is found, \name removes data points from $B$ that belong to the new label. Next, \name uses $B$ to collect new data points with potential new labels. At last, \name builds new prototypes for each new label detected, and updates the parameters of heuristic functions. Each prototype $p=<c,r,\mu,s,\bar{f}>$ is a tuple occupies limited storage space and \name also constrains the maximum number of prototypes in each heuristic function, therefore, \name limits memory usage and computation cost.

%Algorithm 1 depicts the mechanism, including self-adaptation of heuristic functions and corresponding updates on  prototypes. We periodically check the buffer $\mathcal{B}$ and request potential novel label checking. Then, remove data points that do belong to the novel labels. Next, use the buffer $\mathcal{B}$ to receive following data points with potential novel labels. At last, we build new prototypes for each new novel labels detected, and update the parameters for each heuristic function. Because both the max number of prototype of each heuristic function and the storage space for each prototype $p=<c,r,\mu,s,f>$ are both limited. Our system with both memory usage and computation efficiency.

\subsection{Labeling Confidence Aggregator}
\label{sec:labeling_aggregator}

%After the heuristic function $h_i \in H, i \in 1...t$ built, $h_i$ maintains a weight to estimate its confidence for labeling any data point $x \in D_U$, this metric is named as $Conf_i$. Here, we define the $Conf_i$ for $h_i$ as below:

After a heuristic function $h_i$ is built, $h_i$ use a metric to quantify its confidence for labeling any data point $x \in D_U$. This metric, named as $Conf_i$, is defined as follows.

\begin{equation}
\label{eq:confidence}
Conf_i = (r(p_{ik})-d_{ik}(x))\times \frac{\lvert f_{ik}(\ell_{max}) \rvert}{\lvert f_{ik} \rvert}
\end{equation}
Where $p_{ik}$ is the prototype closest to the data point $x$ in the heuristic function $h_i$, $r(p_{ik})$ is the radius of $p_{ik}$, and $d_{ik}(x)$ is the distance between $x$ and $p_{ik}$, $\lvert f_{ik}(\ell_{max}) \rvert$ is the highest label frequency in $f_{ik}$, and $\lvert f_{ik} \rvert$ is the sum of all frequencies in $p_{ik}$. 

In Equation~\ref{eq:confidence}, if $d_{ik}(x)$ is small, which means that $x$ is close to the centroid of $p_{ik}$, then ($r(p_{ik})-d_{ik}(x)$) is large which leads to a high confidence of labeling. In addition, a large $f_{ik}(\ell_{max})$, which means high purity of the prototype $p_ik$, also lead to a high confidence of labeling. Hence, the metric $Conf_i$ considers the impact of both distance and purity, here, purity is calculated by $\frac{\lvert f_{ik}(\ell_{max}) \rvert}{\lvert f_{ik} \rvert}$.
%Thus, here we use this two parts together to represent the confidence of the heuristic function $h_i$. 

\name calculates the confidence value $Conf_i$ for each $h_i$ in $H$. These confidence value are then normalized between 0 and 1, and then aggregated together to calculate the overall labeling confidence of  all heuristic functions, which is shown in Equation~\ref{eq:aggregation}.  

%Finally, each individual heuristic function's confidence is aggregated together to calculate the overall labeling confidence of the entire heuristic functions. 

%Where the closest prototype from data point $x$ in heuristic function $h_i$ is $p_{ik}$, $r(p_{ik})$ is the radius of $p_{ik}$, $d_{ik}(x)$ represents the distance of $x$ from $p_{ik}$, $\lvert f_{ik}(\ell_{max}) \rvert$ is the max label frequency of $x$ in $f_{ik}$, and $\lvert f_{ik} \rvert$ is the sum of all frequencies in $p_{ik}$. The $Conf_i$ is consisted by two parts, if $d_{ik}(x)$ is small which means the $x$ is close to the centroid of $p_{ik}$, then the first part $r(p_{ik})-d_{ik}(x)$ will be large which leads to a high confidence of labeling. For the second part, the large $f_{ik}(\ell_{max})$ means high purity of the prototype $p_ik$, this will also lead to the high $Conf_i$ of $h_i$. Thus, here we use this two parts together to represent the confidence of the heuristic function $h_i$. Our system will calculate the confidence value $Conf_i$ for each $h_i$ in the heuristic function pool, these confidence value are then normalized between 0 and 1. Finally, each individual heuristic function's confidence is aggregated together to calculate the overall labeling confidence of the entire heuristic functions. 
%The final labeling result for $x$ is represented as following.
\begin{equation}
\label{eq:aggregation}
\max_{l \in Y^{'}}\{\sum_{i=1}^{t} {1}(h_i(x)=\ell)\times Conf_i\times h_i(x)\}
\end{equation}

%Here, correlation represents the distance from the data point $x$ to the nearest prototype of $h_i \in H, i \in 1...t$. 
%Therefore, smaller the distance means the higher correlation and higher confidence. %Purity is calculated by the ratio of highest frequency in vector $f$ of the nearest prototype $p=<c,r,\mu,s,f>$ among the length of vector $f$. 
%Hence, higher purity value means higher confidence of each heuristic function. Each heuristic function's correlation and purity value of data point $x$ together will consist its confidence of labeling $x$. 

In Equation~\ref{eq:aggregation} we have a threshold $\tau$ to decide if the confidence is high enough. 
If the overall confidence is higher than $\tau$, the label is assigned; Otherwise, the data point is added into the buffer $\mathcal{B}$ to wait for further checking.

%Here, we will use a threshold $\tau$, if the overall confidence is higher than the $\tau$, the label of the corresponding data point will be assigned, otherwise, the data point will be added into the buffer $\mathcal{B}$ to wait for the further checking. The theoretical guarantees about how confidence value $Conf_i$ improves the labeling result can be found in Appendix.   
\section{Runtime System Design}
\label{sec:sys}

\begin{figure}[tb!]
	\centering
	\includegraphics[width=1.0\linewidth]{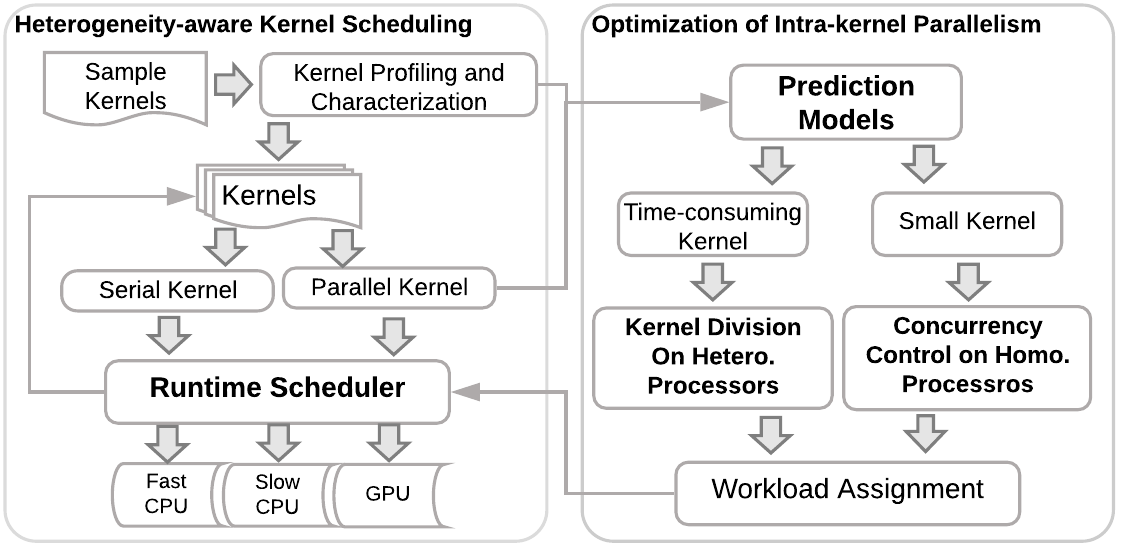}
	\vspace{-10pt}
	\caption{Overview of the runtime system design in \name.} \centering
	\label{fig:sys_overview} 
\end{figure}

%In this section, we demonstrate the heterogeneity-aware mobile runtime system that contains inter- and intra-kernel parallelism to maximize the system throughput and the utilization of heterogeneous mobile processors. The overview of the proposed mobile runtime system is shown in Figure~\ref{fig:sys_overview}. %(The description of the overview figure might be added here if needed.)
We describe the runtime design for \name{} in this section. Figure~\ref{fig:sys_overview} gives an overview of the runtime system. In general, the runtime system includes three techniques: hardware heterogeneity-aware kernel scheduling, concurrency control to determine the number of threads to run a kernel on homogeneous CPU cores, and kernel division on heterogeneous mobile processors. We describe them in details as follows. 

\subsection{Preliminary Performance Study}
\label{sec:prelim_perf}

%\textcolor{dong}{(Describe what are kernels. Describe what are they doing.)}
\name decomposes major computation in the auto-labeling process into kernels. A kernel can be a frequently invoked function; A kernel can also be a computation intensive loop. Table~\ref{tab:kernels} lists kernels in \name. If a kernel has a parallel loop and there is no dependence between iterations of the loop, we name the kernel \textit{parallel kernel}. 
Otherwise, we name the kernel \textit{serial kernel} and run  it with only one CPU thread. %\textcolor{dong}{please verify if the above sentence in red is correct.} 

%In the auto-labeling model, the fine-grained functions each refers to a \textit{kernel}. We refer a kernel as a parallel kernel, if parallel loops in the kernel can be executed in parallel. Otherwise, we refer the kernel as a serial kernel. 

\begin{table}[]
\centering
\caption{A summary of computation kernels in \name.}
\label{tab:kernels}
%\vspace{5pt}
\resizebox{0.48\textwidth}{!}{

\begin{tabular}{|c|c|c|c|}
\hline
Kernels             & \begin{tabular}[c]{@{}c@{}}Parallel or \\ Serial\end{tabular} & Time   & Description                                           \\ \hline
Sample Processing & Parallel                                                      & 38.0\% & Process data samples with heuristic functions      \\ \hline
Detect Change       & Parallel                                                      & 18.0\% & Detect the occurrence of new labels \\ \hline
Test Ensemble       & Parallel                                                      & 17.9\% & Test the ensemble of heuristic functions           \\ \hline
Label Single     & Parallel                                                      & 17.7\% & Label a single instance with heuristic functions        \\ \hline
Warmup Processing   & Serial                                                        & 6.4\%  & Warmup and preparation of \name        \\ \hline
Others (20 in total)            & Parallel \& Serial                                            & 2.0\%  &  Primitive math operations     \\ \hline
\end{tabular}
}

\end{table}

We profile execution times of those kernels using Cifar-10. Table~\ref{tab:kernels} shows the results. We observe that the top five kernels consume more than 95\% of the total execution time in the auto-labeling process. We call the top five kernels the \textit{time-consuming} kernels; The other kernels are the \textit{small} kernels. 
%Given a serial kernel, if its previous kernels need to synchronize before the execution of the serial kernel, this serial kernel is called a \textit{critical-path} kernel. \textcolor{dong}{(the above definition on the critical path is correct???)}
%\textcolor{dong}{(replace the term "rapid kernel" with "small kernel" throughout the paper.)}

%To facilitate the inter- and intra-kernel parallelism in our mobile runtime system design, we characterize kernels in the auto-labeling model as follows. Given a kernel, if the kernel or a portion of the kernel can be executed in parallel, we refer such kernel to \textit{parallel} kernel; otherwise, we refer the kernel to \textit{serial} kernel. We observe that the accumulated execution time of top five kernels in auto-labeling model consume over 90\% of total execution time. Given a parallel kernel, if the kernel belongs to one of the top five kernels, we refer such kernel to \textit{time-consuming} kernel; otherwise, we refer the kernel to \textit{rapid} kernel. Given a serial kernel, if its previous kernels need to synchronize before the execution of the serial kernel, this serial kernel is called a \textit{critical-path} kernel. 

The auto-labeling process using \name can involve a number of iterative steps, typically hundreds or thousands of steps (depending on how many data to be labeled). In each auto-labeling step, a group of data is labeled. We refer a group of data as a \textit{chunk}. At the end of each step, there is a barrier working as a synchronization point where detection of new labels based on a chunk must be finished before \name processes the next chunk. 

The execution time of some kernels highly depends on the number of existing labels. The execution time of those kernels is roughly in linear proportion to the number of existing labels. All parallel kernels in \name are those kernels. Furthermore, for most of the serial kernels, their performance is not related to the number of existing labels, because those serial kernels are used to process input and output data and initialize \name. We leverage the above facts in our analytical models Equations~\ref{eq:Intra_speedup}-~\ref{eq:Intra_division} to optimize execution of individual kernels.

\subsection{Hardware Heterogeneity-aware Kernel Scheduling}
\label{sec:sys_schedule}

\begin{figure}[tb!]
	\centering
	\includegraphics[width=0.9\linewidth]{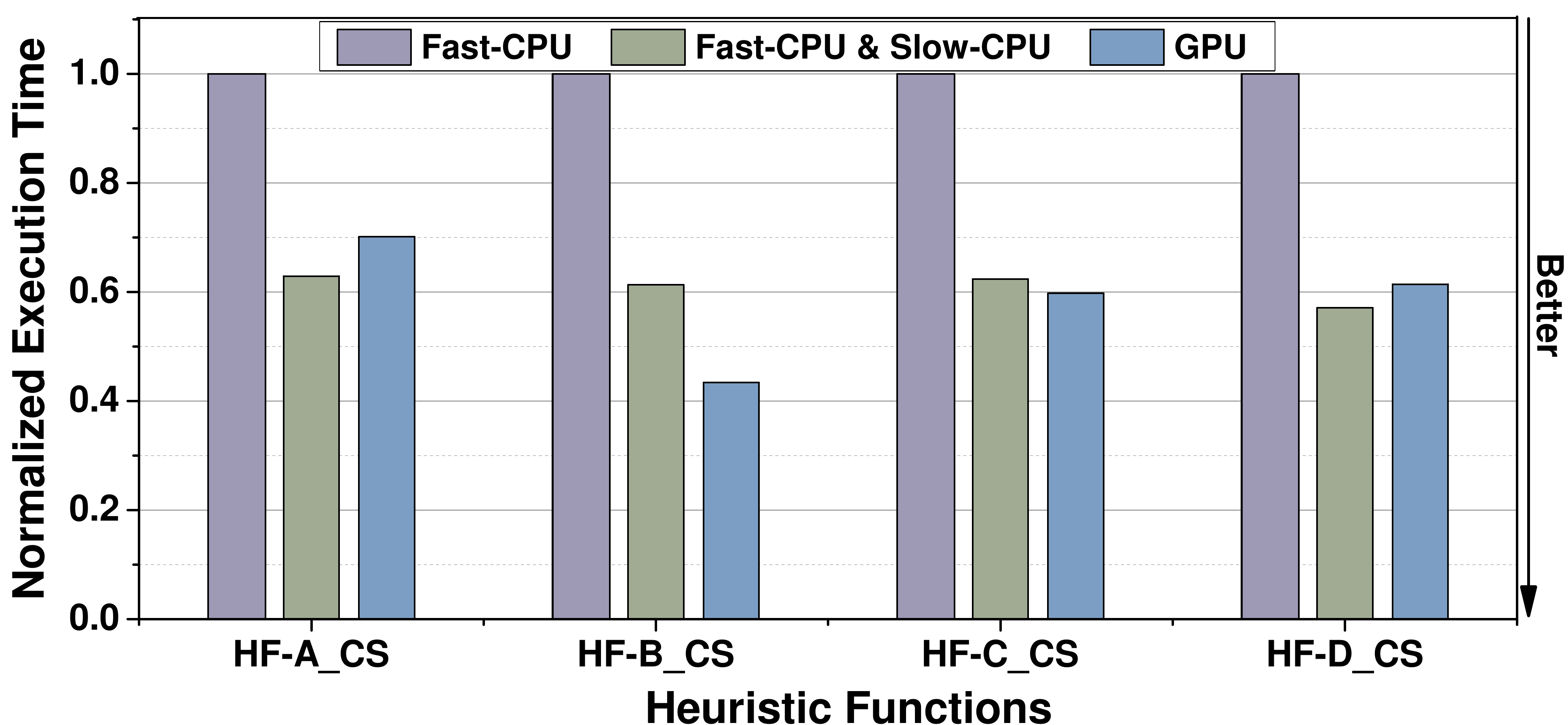}
	%\vspace{-15pt}
	\caption{Normalized execution time for four time-consuming kernels on CPU and GPU. The execution time is normalized by that measured on using fast CPU.}
	%\textcolor{dong}{(in the y axis, use "Normalized execution time". Also, show which direction is better. CD should be updated to CS.)}}
	\vspace{-20pt} 
	\centering
	\label{fig:kernel_perf2} 
\end{figure}

Kernel scheduling has big impact on kernel execution time. To quantify the performance difference of different kernel scheduling, Figure~\ref{fig:kernel_perf2} shows the execution time of four frequently used kernels (HF-A\_CS, HF-B\_CS, HF-C\_CS and HF-D\_CS) running on fast CPU-only, fast and slow CPUs (i.e., using all CPU cores), and GPU. In general, all kernels have performance variance. There is up to 18\% difference between running HF-B\_CS on GPU and all CPU cores.

%Assigning kernels to random mobile processors might have performance degradation and low hardware efficiency. Figure~\ref{fig:kernel_perf2} demonstrates the performance difference between assigning kernels to CPUs and GPU. As shown in Figure~\ref{fig:kernel_perf2}, given kernels HF-A\_CD, HF-B\_CD, HF-C\_CD and HF-D\_CD, if we assign kernels HF-A\_CD and HF-D\_CD to GPU and HF-B\_CD and HF-C\_CD to CPUs, we have 7\% and 5\% performance degradation for kernels HF-A\_CD and HF-D\_CD and 18\% and 3\% performance degradation for kernels HF-B\_CD and HF-C\_CD respectively. Note that kernel HF-A\_CD refers to the kernel of computing distance in one type of heuristic function.

To decide where to run a kernel, we use the following three policies. (1) The time-consuming parallel kernels use all processing units (including both GPU and CPUs), because those kernels are time-consuming and in the critical path. (2) Small and serial kernels use fast CPU, because they cannot benefit from high thread-level parallelism on GPU and can be in the critical path. (3) If all fast CPU cores are busy, we assign kernels to slow CPU cores. 

%To address the mismatch between kernels and mobile processors and optimize the hardware efficiency, we assign a kernel to a mobile processor based on kernel characteristics, hardware features and profiling results and propose following policies: 1) we assign the time-consuming parallel kernels on heterogeneous GPU and CPUs; 2) we assign rapid kernels, critical-path kernels and serial kernels on fast-CPU; 3)if all fast-CPU cores are occupied, we assign kernels to slow-CPU. To assign kernels to mobile processors, we associate a priority queue with each mobile processor. The priority queue has kernels whose dependency have been addressed and ready to run.

The kernel scheduling can suffer from the straggler effect. We describe it as follows. During the auto-labeling process, there is a barrier at the end of each auto-labeling step. The computation of all heuristic functions must finish at the barrier, before the auto-labeling process moves on to the next step. The computation of those heuristic functions happens in parallel. If the computation of one heuristic function finishes much later than the other ones, then we may have idling processing units and hence lose system throughput. Ideally, computation of all heuristic functions should be finished at the same time.

%The kernel scheduling from several heuristic functions at runtime can suffer from the straggler effect. We describe the straggler effect as follows. During the auto-labeling process, there is a barrier at the end of each auto-labeling step to synchronize all concurrent heuristic functions and process new detected labels. The processing of all concurrent heuristic functions must finish at the barrier, before the auto-labeling process moves on to the next auto-labeling step. If the processing of one heuristic function finishes much later than the other heuristic functions, then we lose system throughput and may have low hardware utilization. Ideally, processes of all concurrent heuristic functions should finish at almost the same time.

To address the above problem, we schedule kernels based on the following algorithm. In particular, we associate each kernel with the ID of the heuristic function for which the kernel computes. When the runtime system picks up kernels to execute, the runtime follows a round-robin policy to ensure that kernels from different heuristic functions have the same opportunity to execute. Also, kernels with  long execution time are scheduled to execute first, in order to shorten the critical path of execution, which is also helpful to avoid the straggler effect.

%To address the above problem, we schedule kernels based on the following algorithm. In particular, we associate each kernel with the ID of the heuristic function for which the kernel computes. When the runtime system picks up kernels from the priority queue to run on the mobile processors, runtime follows a round-robin based policy to ensure that kernels from different heuristic have the same opportunity to execute. Also, the kernel in each auto-labeling heuristic function that has the longest execution time will be executed first, to shorten the critical path of execution, which is also helpful to avoid the straggler effect.

%\subsection{Intra-kernel Parallelism}
\subsection{Optimization of Intra-Kernel Parallelism}
\label{sec:sys_intra}

%\subsubsection{Parallelism Granularity} 

%To improve the hardware efficiency of mobile processors, we study and optimize the intra-kernel parallelism on mobile devices. In this section, we focus on the parallelism granularity of mobile CPU. We observe that parallelism granularity on mobile CPU has significant impact for performance of kernels execution.

The execution time of a kernel is sensitive to the number of threads to run it. This is especially true for small kernels running on CPU. Figure~\ref{fig:kernel_perf1} shows the execution time of running four frequently invoked small kernels (HF-A\_DC, HF-B\_DC, HF-C\_DC and HF-D\_DC) with different number of threads. 

%We observe that some rapid kernels of the auto-labeling model on mobile devices are executed frequently and those kernels might not achieve best performance when using all computing elements. Figure~\ref{fig:kernel_perf2} shows the normalized performance variance of some frequently involved rapid kernels as manipulating the number of threads. 

\subsubsection{Concurrency Control} 
\begin{figure}[tb!]
	\centering
	\includegraphics[width=0.9\linewidth]{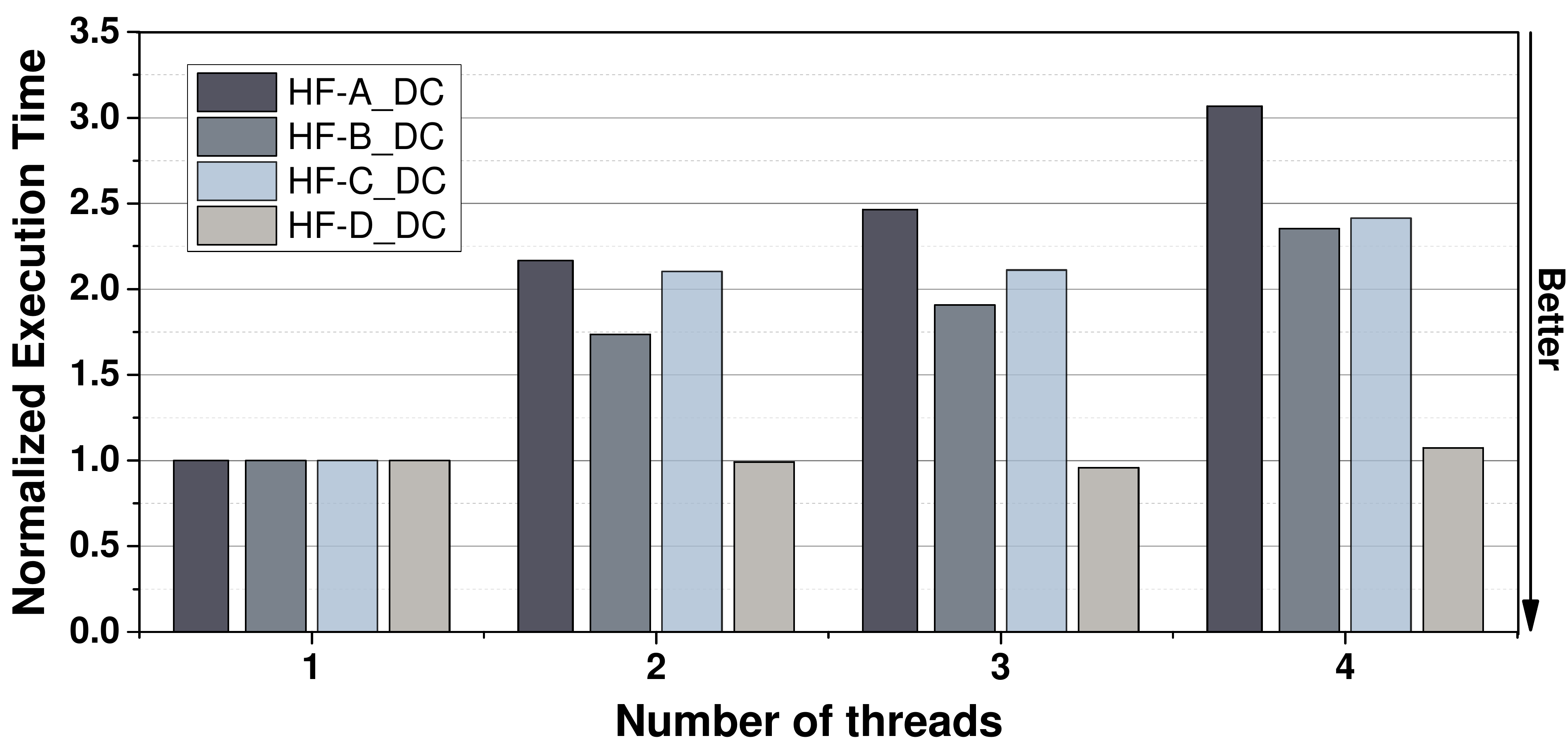}
	\vspace{-12pt}
	\caption{Normalized execution time for four frequently invoked small kernels on fast CPU with different number of threads. The execution time is normalized by that measured using one thread on fast CPU.}
	%\textcolor{dong}{(in the y axis, use "Normalized execution time". Also, show which direction is better. PC should be updated to DC)}}
	\vspace{-10pt}
	\centering
	\label{fig:kernel_perf1} 
\end{figure}

Figure~\ref{fig:kernel_perf1} shows that the kernels HF-A\_DC, HF-B\_DC and HF-C\_DC achieve the best performance using only one CPU thread, while HF-D\_DC achieves the best performance using 3 CPU threads on fast CPU. %\textcolor{dong}{(on fast CPU cores or slow CPU cores?)}. 
The reason accounting for the performance variance is because of thread management overhead (e.g., thread spawning and binding to cores) and cache thrashing due to multi-threading. 
%If we always employ all computing elements to execute a kernel intuitively, we would have performance degradation. For example, we have 3.1X, 2.3X, 2.4X and 11\% performance degradation when executing kernels HF-A\_PF, HF-B\_PF, HF-C\_PF and HF-D\_PF with the maximum number of threads compared to the optimal number of threads. The reason is because of large thread management overhead (e.g., thread spawning or binding to cores) and frequent cache thrashing. Note that HF-A\_PF refers to the kernel of predicting category in one of heuristic function.

To determine the optimal number of threads to run a kernel for best performance, we introduce an analytical model, shown in Equations~\ref{eq:Intra_speedup} and \ref{eq:Intra_parallel}. Table~\ref{tab:model_var} summarizes the notation for the equations. 

%To address the problem, we propose an empirical model to obtain the maximum speedup with available number of threads for the auto-labeling model executed on mobile device. The model is shown in Equation~\ref{eq:Intra_speedup} and Equation~\ref{eq:Intra_parallel}. The notation description is listed in Table~\ref{tab:model_var}.

%\textcolor{dong}{Replace $T_{se}$ and $T_{pe}$ with $T_{ser}$ and $T_{par}$ respectively.}
%\textcolor{dong}{Why do we need to calculate speedup??? Calculating $T_{par}$ is enough.}
%\textcolor{dong}{The equations (5) and (6) are very confusing.}

Equation~\ref{eq:Intra_speedup} calculates performance speedup of using multiple threads to run a kernel. 
In Equation~\ref{eq:Intra_speedup}, $T_{ser}$ is the serial execution time of processing one chunk and there is only one existing label;
$N_{t}$ is the number of threads and $N_{l}$ is the number of labels; $T_{per}$ is the parallel execution time with $N_{t}$ threads and $N_{l}$ labels; $T_{tm}$ is the thread management overhead for one thread; $P$ is the proportion of the execution time in which thread-level parallelism can be employed.

In Equation~\ref{eq:Intra_speedup}, the numerator is the serial execution time of processing $N_l$ labels, where $T_{ser} * (1 - P)$ corresponds to the execution which is not sensitive to the number of labels, and $T_{ser} * P * N_{l}$ corresponds to the execution which is sensitive to the number of labels. The denominator of Equation~\ref{eq:Intra_speedup} is the parallel execution time, including thread management overhead ($T_{tm}*N_t$) and computation time ($T_{per}$) of the parallel kernel. 

Equation~\ref{eq:Intra_parallel} calculates computation time of the parallel kernel, including the serial time ($T_{ser} * (1 - P)$) that is not sensitive to the number of labels and do not run in parallel, and parallel computation time ($\frac{T_{ser}}{N_{t}} * P * N_{l}$).

\begin{table}[]
\centering
\caption{Model parameters for Equations~\ref{eq:Intra_speedup}, ~\ref{eq:Intra_parallel} and ~\ref{eq:Intra_division}}
\label{tab:model_var}
%\vspace{5pt}
\resizebox{0.48\textwidth}{!}{

\begin{tabular}{|l|l|}
\hline
Variable     & Description                                                               \\ \hline
$T_{ser}$     & Serial execution time of processing  one chunk when there is only one existing label                                                 \\ \hline
$N_{t}$      & Number of threads                                                     \\ \hline
$T_{per}$     & Computation time of using $N_{t}$ threads                  \\ \hline
$T_{tm}$     & Thread management overhead for one thread                               \\ \hline
$N_{l}$      & Number of existing labels                               \\ \hline
$P$          & Proportion of the execution time in which thread-level parallelism can be employed \\ \hline
$T_{fast}$      & Execution time on fast CPU                                            \\ \hline
$T_{slow}$      & Execution time on slow CPU                                            \\ \hline
$T_{acc}$      & Execution time on accelerator (GPU)                                   \\ \hline
$T_{ser}^{fast}$ & Serial execution time on fast CPU to process one chunk when there is only one existing label                                   \\ \hline
$T_{ser}^{slow}$ & Serial execution time on slow CPU to process one chunk when there is only one existing label                                    \\ \hline
$T_{exe}^{acc}$     & Execution time on accelerator (GPU) to process one chunk when there is only one existing label                            \\ \hline
$T_{datacpy}^{acc}$     & Time for data copy between CPU and GPU                           \\ \hline
$W_{fast}$      & Percentage of workload assigned to fast CPU                           \\ \hline
$W_{slow}$      & Percentage of workload assigned to slow CPU                           \\ \hline
$W_{acc}$      & Percentage of workload assigned to accelerator (GPU)                  \\ \hline
$N_{t}^{fast}$  & Number of threads to run on fast CPU                                        \\ \hline
$N_{t}^{slow}$  & Number of threads to run on slow CPU                                        \\ \hline
\end{tabular}
}
\vspace{-10pt}
\end{table}

\begin{equation}
\label{eq:Intra_speedup}
\small
Speedup = \frac{T_{ser} * (1 - P) + T_{ser} * P * N_{l}}{T_{tm} * N_{t} + T_{per}}
\end{equation}

\begin{equation}
\label{eq:Intra_parallel}
\small
T_{per} =  T_{ser} * (1-P) + \frac{T_{ser}}{N_{t}} * P * N_{l} 
\end{equation}

%We obtain the values of $T_{se}$, $T_{tm}$ and $P$ in the kernel profiling step and obtain $N_{c}$ at runtime. Since there is a limited number of threads in model CPU, we try the available number of threads in those two equations and compare the calculated speedup with negligible runtime overhead. Thus, using the empirical model, we can obtain the optimal number of threads to execute a kernel.

%\textcolor{dong}{(Please ensure if the following paragraph is correct.)}
%To determine the optimal number of threads to run a kernel, we enumerate various numbers of threads to run kernels to measure $T_{ser}$, $T_{tm}$, $P$, and $N_{c}$. For each number of threads, we use an auto-labeling step. Since the number of threads in a mobile device is limited, we use less than \textcolor{dong}{eight} steps and do not cause large runtime overhead because of the measurement. 
To determine the optimal number of threads to run a parallel kernel, we use the following method. Given a  kernel, $T_{ser}$, $T_{tm}$ and $P$ are measured offline. $N_l$ can be known at runtime.  We enumerate various number of threads and use Equations~\ref{eq:Intra_speedup} and~\ref{eq:Intra_parallel} to find the optimal number of threads that lead to the largest speedup.

%we enumerate various numbers of threads achieve the best speedup. In particular, $T_{ser}$, $T_{tm}$ and $P$ are stable and obtained in the warmup phase and $N_{c}$ is obtained at the runtime. $T_{ser} * (1 - P)$ refers to the serial execution time of kernels which are not sensitive to the number of labels; $T_{ser} * P * N_{l}$ refers to the execution time of parallel kernels which are sensitive to the number of labels using one thread; $\frac{T_{ser}}{N_{t}} * P * N_{l}$ refers to the execution time of parallel kernels which are sensitive to the number of labels using $N_{t}$ thread.

\subsubsection{Heterogeneity-Aware Kernel Division}

%\textcolor{dong}{(the auto-labeling model means heuristic function? if yes, we should use ``heuristic function''.)}

%\textcolor{dong}{kernel division is coupled with the concurrency control, right?}

When running a time-consuming parallel kernel on heterogeneous mobile processors, we must ensure load balance between computing units (GPU, slow CPU and fast CPU). We introduce an analytical model to decide how to divide the computation of a kernel between different computing units (i.e., kernel division) for load balance. The kernel division is implemented by assigning iterations of the loop in a parallel kernel to different computing units. Equation~\ref{eq:Intra_division} shows the model. 

%Kernel division refers to dividing parallel loops within a kernel. The divided workload of the kernel is assigned on different processors individually and the final results are combined. Equation~\ref{eq:Intra_division} shows the model. 

%When dividing a workload among heterogeneous mobile processors, one challenge is to achieve load balance among the mobile processors. To address this challenge, we propose an empirical model processed at runtime to predict optimal workload division of a kernel executing on heterogeneous mobile processors including fast CPU, slow CPU and GPU. The model mainly considers the computational ability of the heterogeneous mobile processors, the profiling results and features of the auto-labeling model. Though our model is applied for heterogeneous CPUs and GPU in the paper, our model can be extended to support other mobile processors, e.g., DSP, NPU, in mobile devices. We leave the model extension to consider more mobile processors as our future work. 

%Though employing a full-fledged performance model, such as XXX (cite some papers using offline prediction models, e.g., NN models, time-consuming models), will provide accurate prediction of the performance, those models also incur large overhead at runtime. Thus, our model is specifically designed for handling time-consuming parallel kernels and are significantly simplified for the needs of parallel kernel division for heterogeneour mobile processors. 

Equation~\ref{eq:Intra_division} is an extension to Equation~\ref{eq:Intra_parallel} which is for homogeneous CPU. Equation~\ref{eq:Intra_division} considers the difference of computation ability in  heterogeneous processors. Given a time-consuming parallel kernel, Equation~\ref{eq:Intra_division} partitions the workload to run on fast CPU ($T_{fast}*W_{fast}$), slow CPU ($T_{slow}*W_{slow}$) and GPU ($T_{acc}*W_{acc}$). Equation~\ref{eq:Intra_division} also considers the computation that is not sensitive to the number of labels ($T_{ser}*(1-P)$) and the computation that is sensitive to the number of labels.  

In Equation~\ref{eq:Intra_division}, given a kernel, $T_{ser}$,  $T_{ser}^{fast}$, $T_{ser}^{slow}$, $T_{exe}^{acc}$, $T_{datacpy}^{acc}$, $T_{fast}$, $T_{slow}$ and $P$ can be measured offline. $N_l$ can be known at runtime. To determine the optimal kernel division, we enumerate the possible values of $N_t^{fast}$, $N_t^{slow}$, $W_{fast}$, $W_{slow}$, and $W_{acc}$, and use Equations~\ref{eq:Intra_division} and~\ref{eq:Intra_speedup} to find the optimal kernel division that leads to the largest speedup.

%The model in Equation~\ref{eq:Intra_parallel} supports the homogeneous CPU. We extend the model to support kernel division for heterogeneous mobile processors in Equation~\ref{eq:Intra_division}. The notation description is listed in Table~\ref{tab:model_var}. In summary, the model considers the computation ability of heterogeneous processors and use profiling results. Given a time-consuming parallel kernel, we divide the workload of the kernel proportional to the parallel execution time of fast CPU ($T_{f}$), slow CPU ($T_{s}$) and GPU ($T_{a}$). We then embed Equation 7 in Equation 5 and dynamically refine $N_{t}^{f}$ and $N_{t}^{s}$ based on the available number of threads in the runtime. After the refinement, we decide the optimal division of $W_{f}$ , $W_{s}$ and $W_{a}$ based on the best speedup we can obtain. Note that the sum of total percentage of a kernel workload, i.e., $W_{f}$ , $W_{s}$ and $W_{a}$, is 100\%.

\begin{align}
\vspace{-25pt}
\!\!\!\!\!\!\!\!\!\!\!\!T_{per} &= T_{fast} * W_{fast} + T_{slow} * W_{slow} + T_{acc} * W_{acc} \nonumber\\
&= T_{ser} * (1-P) + P * N_{l} * (\frac{T_{ser}^{fast} * W_{fast}}{N_{t}^{fast}} + \nonumber\\
&\quad\, \frac{T_{ser}^{slow} * W_{slow}}{N_{t}^{slow}} + (T_{exe}^{acc} + T_{datacpy}^{acc}) * W_{acc})\label{eq:Intra_division} 
 \vspace{25pt}
\end{align}

\section{Implementation}
\label{sec:sys_implementation}
We implement Flame using C++ with Native Development Kit (NDK) on Android 9.0 and evaluate the system on Samsung S9 with Snapdragon 845 SoC. Our implementation includes 3135 lines of code in total. In our mobile platform, we have three types of mobile processors, which are GPU, fast CPU and slow CPU. We implement a scheduler to schedule kernels and divide the computation within a parallel kernel over the three types of mobile processors. To run a kernel on a specific type of CPU, we use the thread affinity API. To execute the kernel on GPU, we maintain a CPU thread to execute an OpenCL version of the kernel. To execute a parallel kernel, we examine the availability of CPU cores and GPU at runtime and then employ the performance models discussed in Section~\ref{sec:sys_intra} to obtain the optimal concurrency and workload division.

\section{Evaluation}
\label{sec:eval}
%The effectiveness of \name primarily is from the quality of auto-labeling results with properly designed model, and efficient model execution on heterogeneous mobile device. In this section, we focus on empirically evaluating the labeling quality and performance of system using several public datasets.
We evaluate \name from the perspectives of labeling quality and execution time on heterogeneous mobile processors.

\subsection{Experimental Setup}

\begin{table}[]
\centering
\caption{Description of datasets.}
\label{tab:data_summary}
%\vspace{5pt}
\resizebox{0.48\textwidth}{!}{

\begin{tabular}{|l|l|l|l|l|l|}
\hline
Type  &   Data Set       & $\#$ of features & $N_{L}$ & $N_{U}$ & $\#$ of labels \\ \hline
Image &   Cifar10        & 3072         & 476 & 4524 & 10        \\ \hline
Image &   MNIST          & 784          & 512 & 4488 & 10        \\ \hline
Image &   Kuzushiji      & 4096         & 432 & 4568 & 10        \\ \hline
Image &   EMNIST         & 3072         & 784 & 4216 & 10        \\ \hline
Detection & SVHN         & 3072         & 500 & 4500 & 10        \\ \hline
Detection & Labelme      & 3072         & 64  & 573  & 8        \\ \hline
Recognition &  SCI       & 4096         & 200 & 1984 & 7        \\ \hline
Recognition &  GTSRB     & 62500        & 500 & 4500 & 40        \\ \hline
\end{tabular}
}

\end{table}

\begin{figure*}[tb!]
	\centering
	\includegraphics[width=1.0\linewidth]{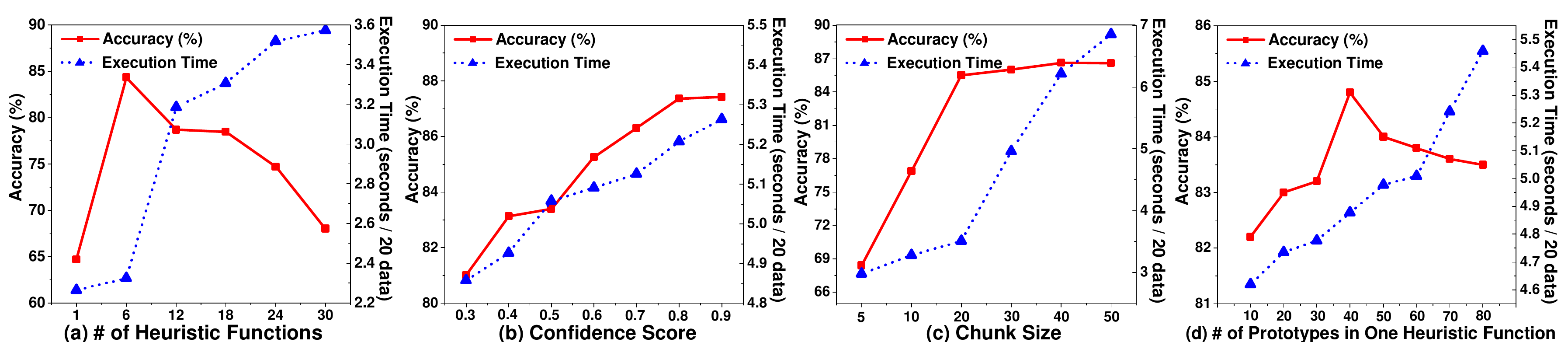}
	\vspace{-10pt}
	\caption{Sensitivity study on how the parameter setting impacts execution time and labeling accuracy.} \centering
	\label{fig:sys_sensitivity} 
	%\vspace{-5pt}
\end{figure*}

\textbf{Datasets.}
We use eight datasets to evaluate \name. Table~\ref{tab:data_summary} summarizes those datasets.
Those datasets are commonly used for object detection or recognition, which are common applications in mobile devices.

%Since our focus is on applications with images, we consider datasets of images for evaluating Flame. Particularly, we use publicly available image datasets for evaluating our system, including MNIST, CIFAR-10, SVHN, LabelMe. The MNIST dataset contains images of articles, each associated with a label from 10 distinct classes. The CIFAR-10 dataset contains a set of natural color images of 32x32 pixels with labels from 10 different classes. The SVHN dataset contains images of house numbers with 32x32 pixels and 3 color channels for each image. Finally, the Labelme dataset is for object detection task, it is created by the MIT Computer Science and Artificial Intelligence Laboratory which provides a dataset of digital images for annotations. Apart from the above popular datasets, we also use four other image datasets that showcase a labeling application, i.e., outdoor scene recognition. Particularly, we use the GTSRB dataset to label for the traffic sign on the road, image sizes vary between 15x15 to 250x250 pixels. The EMNIST dataset containing enlarged MNIST images. Another dataset called Style Color Images consisted by photos sorted by products and brands, each image contains 150x150 pixels and 3 channels. Besides these datasets, we also evaluate our system on Kuzushiji, the dataset is consisted by Kanji characters, each image contains 64x64 pixels. Details of all the above datasets are listed in Table.

\name can detect unknown new labels, when new data are dynamically generated in a mobile device. To evaluate this ability of \name, we use the following method. 
For each dataset, we choose 20\% of labels as known, and the rest of labels as unknown. The rest of labels needs to be detected by \name. We choose a subset of data $D_{L}$ from each dataset. The data subset has known labels, and is used to build heuristic functions at the beginning of auto-labeling. Excluding $D_{L}$, the rest of dataset ($D_{U}$) is used to simulate the scenario where new data is incrementally generated for auto-labeling. The ratio between $D_{L}$ and $D_{U}$ is 0.1. 

We use six heuristic functions, each of which contains 40 prototypes. The dynamically generated data is fed into \name in the granularity of chunk. A chunk includes 20 data samples. The labeling confidence threshold $\tau$ is set as 0.7. 
%We evaluate the impact of parameter setting in our study.

%When users use the mobile devices, some new generated data with novel labels may occur. Here, we briefly illustrates how to simulate this process through the public datasets. This paper only focuses on applications with images, for these selected datasets, we rearrange the images in each dataset to emulate the process about dynamically incremental data with novel labels, i.e., new generated images without labels occurring sequentially over a non-stationary setting. Some of these datasets used in this paper with too many images, e.g, CIFAR-10 contains 60000 images. It is rare for mobile devices to generate such large number of images during the usage. Therefore, we just select at most 5000 images for those datasets with over 5000 images. The novel labels induced as follows. At the beginning, we group images in each public dataset according to their labels, consider 20\% of those labels are known and the rest of them are novel. We then select the limited number of images with labels for building initial heuristic functions from known labels at random. Initially, this pool consists of only images from known labels, denoted as $D_{L}$. At last, we randomly select a large number of images without labels from the dataset excludes $D_{L}$ to consist unlabeled dataset, denoted as $D_{U}$. The ratio about $D_{L}$ and $D_{U}$ is around 10\% in the paper.  

\textbf{Mobile device configuration.}
%We evaluate the accuracy, performance and energy efficiency of Flame on Samsung S9 equipped with Snapdragon 845 SoC and Android 9.0 Pie OS. In Snapdragon 845 SoC, we use Adreno 630 GPU which we program with OpenCL 2.0~\cite{opencl}. To evaluate Flame, we use 6 heuristic functions, each heuristic function contains 40 prototypes, the dynamic generated data flowed into the system as a data chunk, we set chunk size as 20 in the paper. Furthermore, the threshold for labeling confidence is 0.7. The sensitivity of these parameters are studied in the parameter sensitivity section.  
We evaluate \name on a Samsung S9 smartphone. This device is equipped with Snapdragon 845 SoC and Android 9.0 Pie OS. In Snapdragon 845 SoC, there is a mobile GPU, Adreno 630. We program it with OpenCL 2.0~\cite{opencl}. 

%\subsection{Evaluation Metrics}
\textbf{Evaluation metrics.}
We use the following metrics to evaluate the system's labeling results.

$Accuracy\%: \frac{N_{new}+N_{exist}}{N}$, where $N_{new}$ is total number of data with new labels correctly labeled, 
$N_{exist}$ is the number of data with existing labels identified correctly, and $N$ is total number of data labeled by the system.

Let $FP$ represent the total data that should be assigned with existing labels but is mislabeled with new labels (i.e., previously unknown labels); Let $FN$ represent the total data that should be assigned with new labels but is mislabeled with existing labels; Let $N_{l}$ the total number of data assigned with new labels. To evaluate labeling results, besides $Accuracy\%$, we use another three metrics based on $FP$, $FN$, and $N_{l}$. 

$M_{new}$: Percentage of data that should be assigned with new labels but is mislabeled with existing labels, i.e. $\frac{FN∗100}{N_{l}}$. 

$F_{new}$: Percentage of data that should be assigned with existing labels but is mislabeled with new labels, i.e. 
$\frac{FP∗100}{N-N_{l}}$. 

$F_{\beta}$: This metric quantifies the overall labeling quality of the system by considering both $precision$ and $recall$. $F_{\beta}$ is defined as $F_{\beta} = \frac{(1+\beta^{2})*TP}{(1+\beta^{2})*TP+\beta^{2}*FN+FP}$, where $TP$ is the total number of data that should be assigned with new labels and assigned correctly. In this paper, we use $\beta = 2$, which gives $F_{2} = \frac{5*TP}{5*TP+4*FN+FP}$.

\begin{table}[]
\centering
\caption{Summary of labeling results.}
\label{tab:model_results}
%\vspace{5pt}
\resizebox{0.43\textwidth}{!}{
\begin{tabular}{|l|l|l|l|l|}
\hline
Data Set     & Accuracy\% & $M_{new}$ & $F_{new}$ & $F_{2}$                                                               \\ \hline
Cifar10     & 72.45  & 26.57 & 9.27 & 0.76                      \\ \hline
MNIST       & 87.42  & 11.08 & 12.76 & 0.89                      \\ \hline
Kuzushiji         & 82.85  & 17.30 & 5.92 & 0.85                    \\ \hline
EMNIST         & 78.14  & 16.98 & 22.02 & 0.84                     \\ \hline
SVHN       & 71.59  & 37.84 & 5.92 & 0.73                     \\ \hline
Labelme       & 83.72  & 18.05 & 6.63 & 0.84                     \\ \hline
SCI        & 74.76  & 26.88 & 4.47 & 0.77                     \\ \hline
GTSRB       & 68.53  & 31.66 & 8.67 & 0.72                     \\ \hline
\end{tabular}}
\vspace{-10pt}
\end{table}

\subsection{Evaluation Results}
\textbf{Labeling results.} 
Table~\ref{tab:model_results} shows the results, evaluated with the eight datasets based on the four different evaluation metrics. In general, \name provides good labeling quality, This labeling quality is comparable to that in the existing work \cite{varma2018snuba, ratner2017snorkel}.

The labeling quality on MNIST is the best: $Accuracy\% = 87.42$ and $F_{2} = 0.89$. They are the highest among all datasets. This is because the image characteristics with different labels in MNIST are quite dissimilar, making the work of auto-labeling easier. However, for the GTSRB dataset, \name has a relatively low labeling quality ($Accuracy\%=68.53$). This is because the data in this dataset comes from 40 different classes and using our simulation method to simulate unknown labels, we have 32 unknown labels. Such a large number of unknown labels can influence the labeling quality. For  $M_{new}$, the results of this metric show that \name has a good ability to detect new labels as dataset dynamically increases, especially for the dataset related with image classification. This is due to the effectiveness of the self-adaptation mechanism in \name, which is superior to other auto-labeling methods that can only work for a fixed size of datasets. Besides the superior ability to detect new labels, \name also maintains a high labeling quality for the data that should be assigned with the existing labels. This fact is supported by the results of $F_{new}$, where $F_{new}$ can be as small as 4.47.

\begin{figure*}[tb!]
	\centering
	\includegraphics[width=\linewidth]{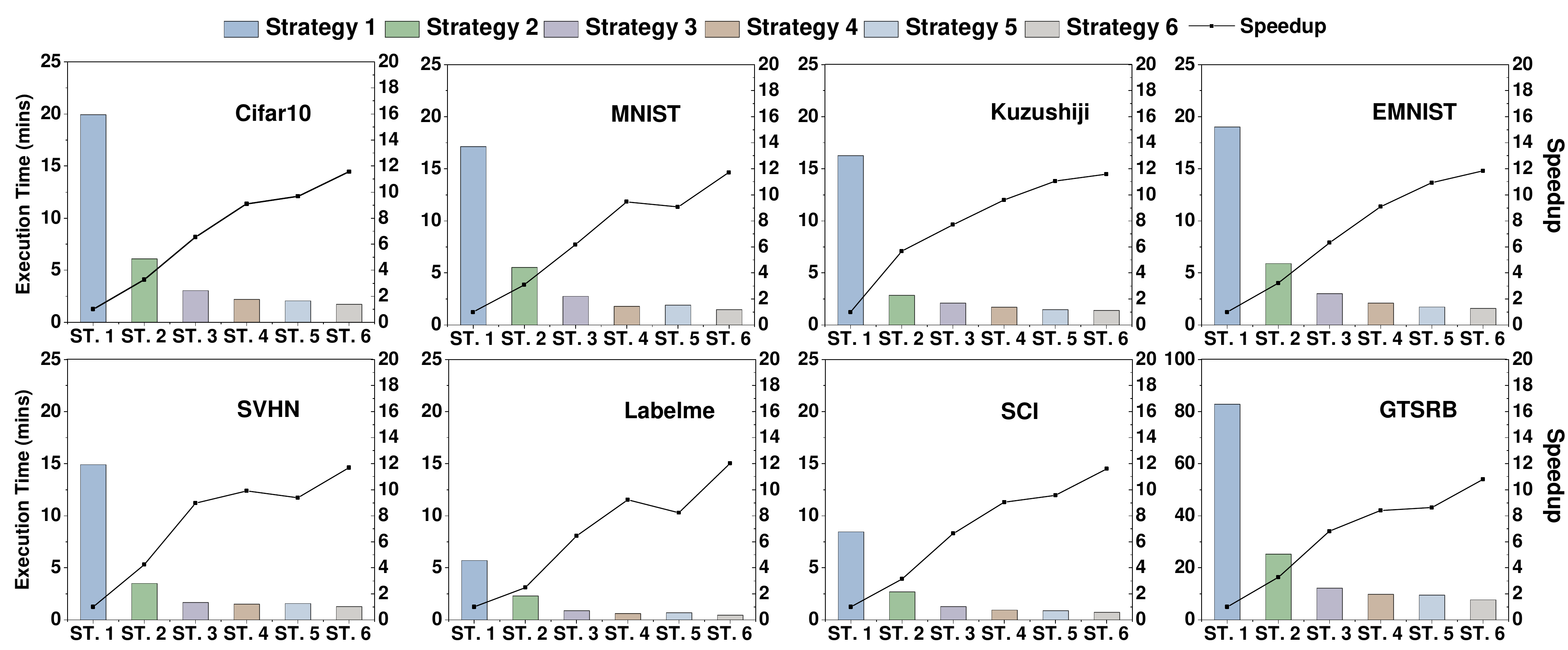}
	%\vspace{-20pt}
	\caption{Execution time and performance speedup for eight datasets with six strategies. ST.1 : serial execution with only one fast CPU core with a FIFO scheduling strategy; ST.2: parallel execution using CPUs (fast and slow CPU cores) with the FIFO scheduling strategy; ST.3: parallel execution using CPUs and GPU with the FIFO scheduling strategy; ST.4: parallel execution using CPUs and GPU with optimization on kernel scheduling but not on concurrency control and heterogeneity-aware kernel division; ST.5: parallel execution using CPUs and GPU with the FIFO scheduling strategy and heterogeneity-aware kernel division; ST.6: parallel execution using all techniques including kernel scheduling, concurrency control and heterogeneity-aware kernel division. We use the first strategy as the baseline for comparison.} 
	%\vspace{-5pt}
	\centering
	\label{fig:performance} 
\end{figure*}

%As shown in Table~\ref{tab:model_results}, our system has decent labeling quality among the eight datasets based on 4 different evaluation metrics. The labeling quality on the MNIST is the best, due to the characteristics of the images with different labels in this data are dissimilar. Meanwhile, on GTSRB dataset, our system has the relatively low labeling quality. Because the data in this dataset are from 40 different classes, the large number of new coming labels can influence the labeling quality. The results of $M_{new}$ value show our system has good ability to detect the new coming labels as dataset dynamically increases, especially for the application related with image classification. This is due to the labeling heuristics self-adaptation mechanism in the system, which is superior to other auto-labeling methods that just work for fixed size of datasets. Besides the superior ability to detect the new labels, our system also maintains a high labeling quality for those coming data are from existing labels as the value of $F_{new}$ metric shows.  

% Figure~\ref{fig:energy_efficiency
\textbf{Parameter sensitivity.}
We evaluate how execution time and labeling accuracy ($Accuracy\%$) vary, as we use different system configurations (particularly the number of heuristic functions, the confidence score threshold $\tau$, chunk size, and the number of prototypes in a heuristic function). 
Figure~\ref{fig:sys_sensitivity} shows the results. 

Figure~\ref{fig:sys_sensitivity}.a shows that the accuracy increases as the number of heuristic functions increases but is smaller than 10. However, the accuracy drops down when the number of heuristic functions is larger than 10. This is because too many heuristic functions cause an overfitting problem. Furthermore, the execution time consistently increases as the number of heuristic functions increases. This is expected, because the execution time of \name is related to the number of heuristic functions. 

%Figure~\ref{fig:sys_sensitivity} indicates that the accuracy increases with the number of heuristic functions at the beginning. However, the accuracy then decreases as too many heuristic function may cause overfitting problem. Meanwhile, the execution time also increases with increasing number of heuristic functions. This is expected since the execution time of Flame is related to the number of heuristic functions. 

Figure~\ref{fig:sys_sensitivity}.b shows that the accuracy increases as the confidence score threshold increases. However, the execution time dramatically increases after the confidence score threshold is larger than 0.7. Hence, we use $\tau = 0.7$ in \name.

Figure~\ref{fig:sys_sensitivity}.c shows that the accuracy increases as the chunk size increases. This is because as the chunk size increases, the diversity of the data within the chunk also increases. The high data diversity influences the labeling quality. However, if the chunk size is larger than 20, increasing the chunk size is not helpful to improve the accuracy.  Also, the execution time increases a lot as the chunk size is larger than 20. Hence, we choose 20 as the chunk size in \name.

Figure~\ref{fig:sys_sensitivity}.d shows that the accuracy increases as the number of prototypes in a heuristic function increases. But if the number of prototypes is larger than 40, increasing the number of prototypes is not helpful for increasing accuracy. This is because too many prototypes in a heuristic function may cause an overfitting problem for labeling results. Hence, we choose 40 as the number of prototypes in \name.

%Figure 6 demonstrates the accuracy increases with the value of confidence score threshold, however, the execution time dramatically increases after confidence score threshold $\tau = 0.7$, thus, we choose $\tau = 0.7$. In our system, several chunks of data are parallelly processed, the accuracy increases as the increasing chunk size, then after $chunk size = 20$, accuracy does not synchronously increase with the chunk size, but the execution time increases a lot. Therefore, we choose chunk size as 20 in our system. Furthermore, the number of prototypes in one heuristic function also affects the accuracy and execution time of the system. When the number of prototypes in one heuristic function is 40, the accuracy achieves the highest value. After that, the accuracy decreases.

%\subsection{Execution Time}
\textbf{Execution time.}
Figure~\ref{fig:performance} presents the execution time of labeling 5000 data samples over eight datasets. We use six execution strategies to evaluate the effectiveness of \name. These strategies are (1) serial execution with only one fast CPU core with a FIFO scheduling strategy; (2) parallel execution using CPUs (fast and slow CPU cores) with the FIFO scheduling strategy; (3) parallel execution using CPUs and GPU with the FIFO scheduling strategy; (4) parallel execution using CPUs and GPU with optimization on kernel scheduling but not on concurrency control and heterogeneity-aware kernel division; (5) parallel execution using CPUs and GPU with the FIFO scheduling strategy and heterogeneity-aware kernel division; (6) parallel execution using all techniques including kernel scheduling, concurrency control and heterogeneity-aware kernel division. We use the first strategy as the baseline for comparison.

%To demonstrate the effectiveness of our runtime system, we study the  performance of employing Flame on heterogeneous mobile device. Figure~\ref{fig:performance} presents the execution time of labeling 5000 data samples over eight datasets and the speedup across six execution strategies. The six execution execution include 1) default execution without parallelism; 2) parallel execution using all CPUs (fast and slow GPUs); 3) parallel execution using all CPUs and GPU with the default scheduling strategy (FIFO); 4) parallel execution using all CPUs and GPU with inter-kernel parallelism 5) parallel execution using all CPUs and GPU with the default scheduling strategy and the heterogeneity-aware intra-kernel parallelism; 6) putting all together. We use strategy 1 as the baseline in the evaluation.

Figure~\ref{fig:performance} shows 3.6x, 6.9x, 9.2x, 9.6x and 11.6x performance improvement on average after applying strategies 2-6 respectively. \name (Strategy 6) leads to be largest performance improvement.  Using Strategy 2, each kernel is executed by leveraging multiple CPU cores, leading to 3.6x performance improvement. However, GPU is idling. In Strategy 3, both CPUs and GPU are utilized. As a result, the performance speedup is increased to 6.9x. However, the kernel scheduling (FIFO) is not efficient. Strategy 4 improves the kernel scheduling by considering hardware heterogeneity. The performance speedup is increased from 6.9x to 9.2x. Strategy 5 does not uses the kernel scheduling in \name, but uses the heterogeneity-aware kernel division. This optimization also leads to big performance improvement (9.6x). But Strategy 6 after using all techniques lead to the largest performance improvement.

\begin{figure}[tb!]
	\centering
	\includegraphics[width=0.8\linewidth]{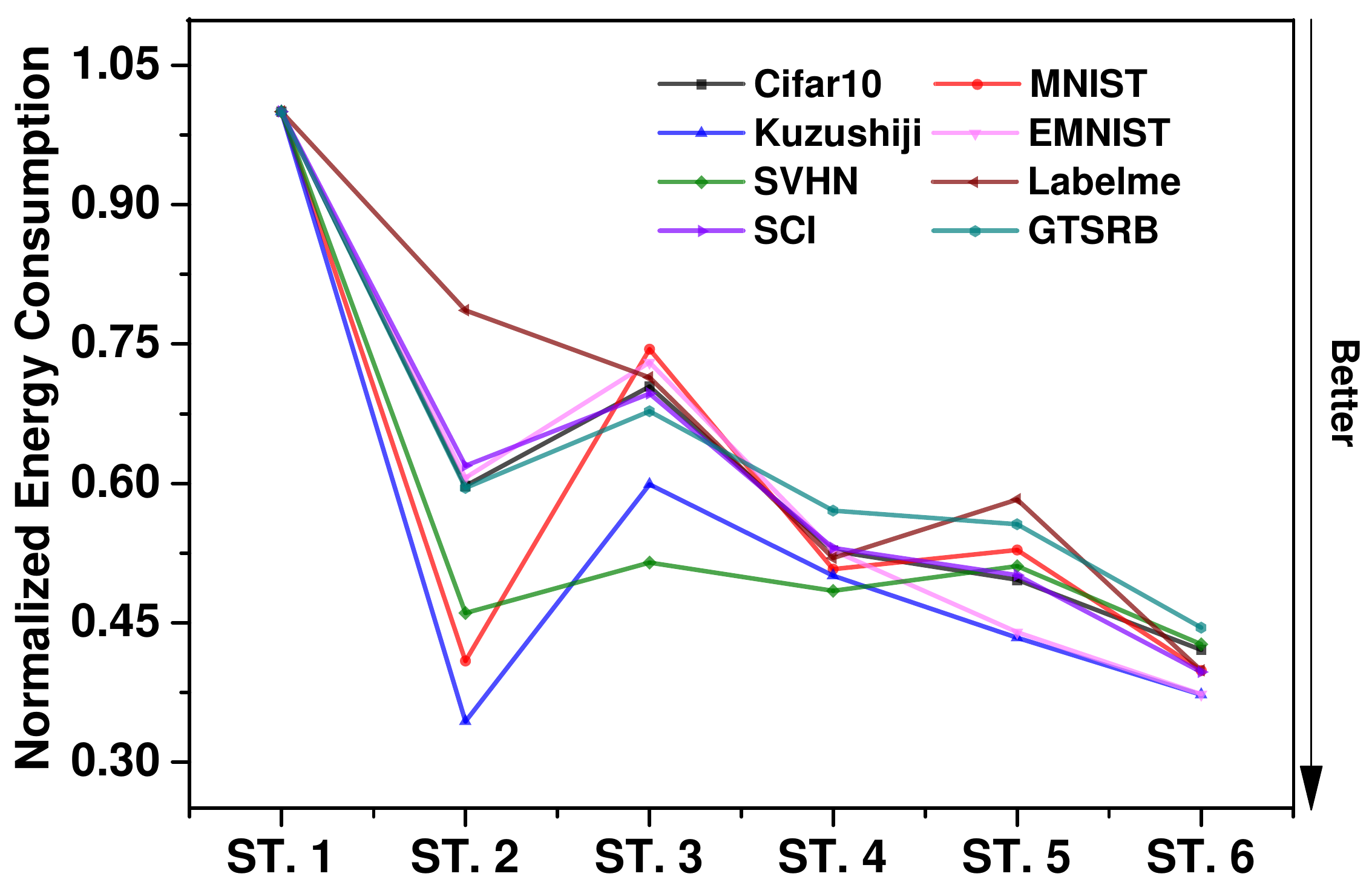}
	\vspace{-5pt}
	\caption{Energy consumption of Flame using eight datasets and six strategies.} \centering
	\vspace{-12pt}
	\label{fig:energy_efficiency} 
\end{figure}

%\subsection{Energy Consumption}
\textbf{Energy consumption.} 
Figure~\ref{fig:energy_efficiency} shows the energy consumption of six strategies. The energy consumption reported in Figure~\ref{fig:energy_efficiency} is normalized by the energy consumption of Strategy 1. Figure~\ref{fig:energy_efficiency} shows that energy consumption of Strategies 2-6 is 55\%, 67\%, 52\%, 51\% and 40\% of that of Strategy 1, respectively. \name (Strategy 6) uses the least energy. Having low energy consumption is important for mobile devices to extend its lifetime. Low energy consumption of \name comes from its high performance, i.e., labeling data within the shortest time among all strategies. 
%Although Strategy 2 with dataset Kuzushiji slightly saves more energy than Strategy 6 (3\%) because of no use of GPU, Strategy 6 leads to 2.1x performance improvement than Strategy 2 for Kuzushiji. 

%To show that Flame is energy-efficient for heterogeneous mobile processors, we study the energy efficiency of Flame. Figure~\ref{fig:energy_efficiency} shows the normalized energy efficiency of Flame across eight datasets with six strategies. We profile the power consumption of mobile device by Trepn~\cite{trepn}. We use energy-delay-product (EDP) as the metric to evaluate energy efficiency. The energy efficiency reported in the figure is normalized by the energy efficiency of applying Strategy 1. In Figure~\ref{fig:energy_efficiency}, the lower the values are, the more energy-efficient. Compared to the Strategy 1, Flame is much more energy-efficient. As shown in the figure, EDP of Flame are 5.9x, 10.1x, 17.6x, 18.6x and 28.7x less than Strategy 1 on average for Strategy 2, 3, 4, 5 and 6 respectively. Also, EDP of Strategy 4, 5, 6 are 1.8x, 1.9x and 2.9x less than Strategy 3.
\section{Related Work}
\label{sec:related}
\textbf{Automatic labeling.} We provide an overview of automatic labeling methods, which label data automatically based on generated heuristic functions using both labeled and unlabeled data.

The main challenge for auto-labeling is to build proper heuristic functions that can cover almost all data in the dataset~\cite{wang2015falling, wang2015or, Varma2018SnubaAW, Varma2017InferringGM, ratner2016data, bach2017learning}. Heuristic functions with high quality are usually difficult to acquire and can be highly application specific. Sometimes domain experts are even needed for auto-labeling. In~\cite{Varma2018SnubaAW}, Varma et. al propose a method that uses machine learning models to build heuristic functions under weak supervision. Other work~\cite{hastie2009multi, weiss2016survey, ratner2017snorkel} uses distant supervision~\cite{hastie2009multi, weiss2016survey}, in which the training sets are generated with the help of external resources, such as knowledge bases. This kind of method called crowdsource has been intensively studied~\cite{li2017human, chai2016cost, khan2016attribute, verroios2017waldo, das2017falcon, wang2012crowder}, and has been applied in many fields, such as task generation~\cite{wang2012crowder}, image labeling~\cite{ratner2017snorkel, Varma2018SnubaAW} and task selection~\cite{verroios2017waldo}.

Some approaches are recently proposed for noisy or weak heuristic functions~\cite{sheng2008get, ratner2017snorkel}. Those approaches demonstrate the use of proper strategies to boost the overall quality of labeling by ensemble heuristic functions~\cite{bach2017learning}. Our work is different from those approaches. The existing approaches focus on static datasets with fixed size and pre-determined number of labels, and datasets are deployed on a server. Our work focuses on the dynamically increased datasets on mobile devices. Our work has the capability to identify new labels that are never seen before. We also leverage processor heterogeneity in mobile devices to run the auto-labeling workload. Hence, our work not only labels dataset with high quality, but also is specific for mobile devices. 
%to improve performance superiority that is shown in our experiments.

\textbf{Optimization of machine learning on mobile devices.}
There are many existing efforts that optimize machine learning models on mobile devices, including dynamic resource scheduling~\cite{georgiev2016leo, likamwa2015starfish,lane2016deepx,liu2019runtime, ogden2018modi}, computation pruning~\cite{gordon2018morphnet,li2018deeprebirth}, model partitioning~\cite{Kang:2017:NCI:3037697.3037698, lane2016deepx}, model compression~\cite{fang2018nestdnn, liu2018demand}, coordination with cloud servers~\cite{georgiev2016leo, Kang:2017:NCI:3037697.3037698} and memory management~\cite{fang2018nestdnn, likamwa2015starfish}. In particular, DeepX~\cite{lane2016deepx} proposes a number of resource scheduling algorithms to decompose DNNs into different sub-tasks on mobile devices. LEO~\cite{georgiev2016leo} introduces a power-priority resource scheduler to maximize energy efficiency. NestDNN~\cite{fang2018nestdnn} compresses and prunes models based on the available hardware resource on mobile devices. Our work is different from those efforts, in that we introduce an efficient hardware heterogeneity-aware kernel scheduling and focus on optimization of intra-kernel parallelism to achieve high performance and energy consumption.
\section{Conclusions}
\label{sec:conclusion}
Auto-labeling on mobile devices is critical to enable successful ML training on mobile devices for many large ML models. However, it is challenging to enable auto-labeling on mobile devices, because of unique data characteristics on mobile devices and heterogeneity of mobile processors. In this paper, we introduce the first auto-labeling system named \name to address the above problem. \name includes an auto-labeling algorithm to detect new unknown labels from non-stationary data; It also includes a runtime system that efficiently schedules and executes auto-labeling workloads on heterogeneous mobile processors. Evaluating with eight datasets, we demonstrate that \name enables auto-labeling with high labeling accuracy and high performance.

\nocite{langley00}

\bibliography{main}
\bibliographystyle{sysml2019}

%%%%%%%%%%%%%%%%%%%%%%%%%%%%%%%%%%%%%%%%%%%%%%%%%%%%%%%%%%%%%%%%%%%%%%%%%%%%%%%
%%%%%%%%%%%%%%%%%%%%%%%%%%%%%%%%%%%%%%%%%%%%%%%%%%%%%%%%%%%%%%%%%%%%%%%%%%%%%%%
% SUPPLEMENTAL CONTENT AS APPENDIX AFTER REFERENCES
%%%%%%%%%%%%%%%%%%%%%%%%%%%%%%%%%%%%%%%%%%%%%%%%%%%%%%%%%%%%%%%%%%%%%%%%%%%%%%%
%%%%%%%%%%%%%%%%%%%%%%%%%%%%%%%%%%%%%%%%%%%%%%%%%%%%%%%%%%%%%%%%%%%%%%%%%%%%%%%
%\appendix
%\section{Please add supplemental material as appendix here}
%
%Put anything that you might normally include after the references as an appendix here, {\it not in a separate supplementary file}. Upload your final camera-ready as a single pdf, including all appendices.

%%%%%%%%%%%%%%%%%%%%%%%%%%%%%%%%%%%%%%%%%%%%%%%%%%%%%%%%%%%%%%%%%%%%%%%%%%%%%%%
%%%%%%%%%%%%%%%%%%%%%%%%%%%%%%%%%%%%%%%%%%%%%%%%%%%%%%%%%%%%%%%%%%%%%%%%%%%%%%%

\end{document}